%% file: colm2025_conference.tex
\documentclass{article}

\usepackage[final]{colm2025_conference}

\usepackage{microtype}
\usepackage{tabularx}
\usepackage{hyperref}
\usepackage{url}
\usepackage{booktabs}
\usepackage{graphicx}
\usepackage{lineno}
\usepackage{color}
\usepackage{amssymb}
\usepackage{paralist}

\usepackage{url}
\usepackage{breakurl}  
\usepackage{hyperref}

\usepackage{subcaption}
\usepackage{graphicx}
\usepackage{latexsym}
\usepackage{amsmath}
\usepackage{float}
\usepackage{footnote}
\usepackage{enumitem}
\usepackage{bm}
\usepackage{arydshln}
\usepackage{booktabs}
\usepackage{multicol}
\usepackage{multirow}
\usepackage{color} 
\usepackage{colortbl}
\usepackage{bbding}
\usepackage{makecell}
\usepackage{mathtools}
\usepackage{imakeidx}
\usepackage{longtable}
\usepackage{wrapfig}
\usepackage{rotating}

\usepackage{arydshln}
\usepackage{lipsum}
\usepackage{natbib}
\usepackage[toc]{multitoc}
\usepackage[edges]{forest}
\usepackage[normalem]{ulem}
\definecolor{mydarkblue}{rgb}{0,0.08,0.45}
\usepackage{CJKutf8}
\usepackage{awesomebox} 
\usepackage{bbding}
\usepackage[most]{tcolorbox}
\usepackage{booktabs}

\definecolor{wkblue}{rgb}{0.2, 0.3, 0.6}
\definecolor{meta-color}{rgb}{0.5, 0.5, 0.5}
\usepackage{amsmath}
\usepackage{enumitem}
\usepackage{lscape} 
\usepackage{booktabs}
\usepackage{algorithm}      
\usepackage{algorithmicx}
\usepackage{algpseudocode}
\usepackage{tabularx,booktabs}
\usepackage{makecell}

\usepackage{amsfonts}

\definecolor{darkblue}{rgb}{0, 0, 0.5}
\hypersetup{
    colorlinks=true,
    citecolor=darkblue,
    linkcolor=darkblue,
    urlcolor=darkblue
}

\DeclareCaptionFont{black}{\color{black}}

\definecolor{myblue}{rgb}{0.9, 0.1, 0.94}
\definecolor{mygreen}{rgb}{0.64, 0.56, 0.88}
\definecolor{myyellow}{rgb}{0.68, 0.6, 0.1}
\definecolor{fancygreen}{rgb}{0.33, 0.68, 0.20}
\definecolor{salmon}{rgb}{0.94, 0.52, 0.49}
\definecolor{tablegreen}{rgb}{0.82, 0.94, 0.75}
\definecolor{tableblue}{rgb}{0.81, 0.90, 0.94}
\definecolor{tablered}{rgb}{0.97, 0.85, 0.85}
\definecolor{tableorange}{rgb}{0.96, 0.85, 0.81}

\newenvironment{itemize*}%
 {\leftmargini=10pt\begin{itemize}%
  \setlength{\itemsep}{0pt}%
  \setlength{\parskip}{0pt}%
  }%
 {\end{itemize}}
\newenvironment{enumerate*}%
 {\begin{enumerate}%
  \setlength{\itemsep}{0pt}%
  \setlength{\parskip}{0pt}}%
 {\end{enumerate}}

\title{LIMO: Less is More for Reasoning}

\author{Yixin Ye$^{1,2}$\thanks{Co-first authors}, Zhen Huang$^{2,3}$\footnotemark[1], Yang Xiao$^{4}$, Ethan Chern$^{1,2}$, Shijie Xia$^{1,2}$, Pengfei Liu$^{1,2}$\thanks{Corresponding author}
\\
$^1$Shanghai Jiao Tong University \quad $^2$SII-GAIR \\
$^3$Fudan University \quad
$^4$The Hong Kong Polytechnic University 
}

\begin{document}
    \ifcolmsubmission \linenumbers \fi

    \maketitle

    \begin{abstract}
        We challenge the prevailing assumption that complex reasoning in large language models (LLMs) necessitates massive training data. We demonstrate that sophisticated mathematical reasoning can emerge with only a few examples. 
        Specifically, through simple supervised fine-tuning, our model, LIMO, achieves 63.3\% accuracy on AIME24 and 95.6\% on MATH500, surpassing previous fine-tuned models (6.5\% on AIME24, 59.2\% on MATH500) while using only 1\% of the training data required by prior approaches. Furthermore, LIMO exhibits strong out-of-distribution generalization, achieving a 45.8\% absolute improvement across diverse benchmarks, outperforming models trained on 100× more data.
        Synthesizing these findings, we propose the Less-Is-More Reasoning Hypothesis (LIMO Hypothesis): In foundation models where domain knowledge has been comprehensively encoded during pre-training, sophisticated reasoning can emerge through minimal but strategically designed demonstrations of cognitive processes. This hypothesis suggests that the threshold for eliciting complex reasoning is not dictated by task complexity but rather by two key factors: (1) the completeness of the model's pre-trained knowledge base and (2) the effectiveness of post-training examples in serving as “cognitive templates” that guide reasoning.\footnote{\url{https://github.com/GAIR-NLP/LIMO}}

    \end{abstract}

    \begin{figure}[ht]
        \centering
        \includegraphics[width=0.8\linewidth]{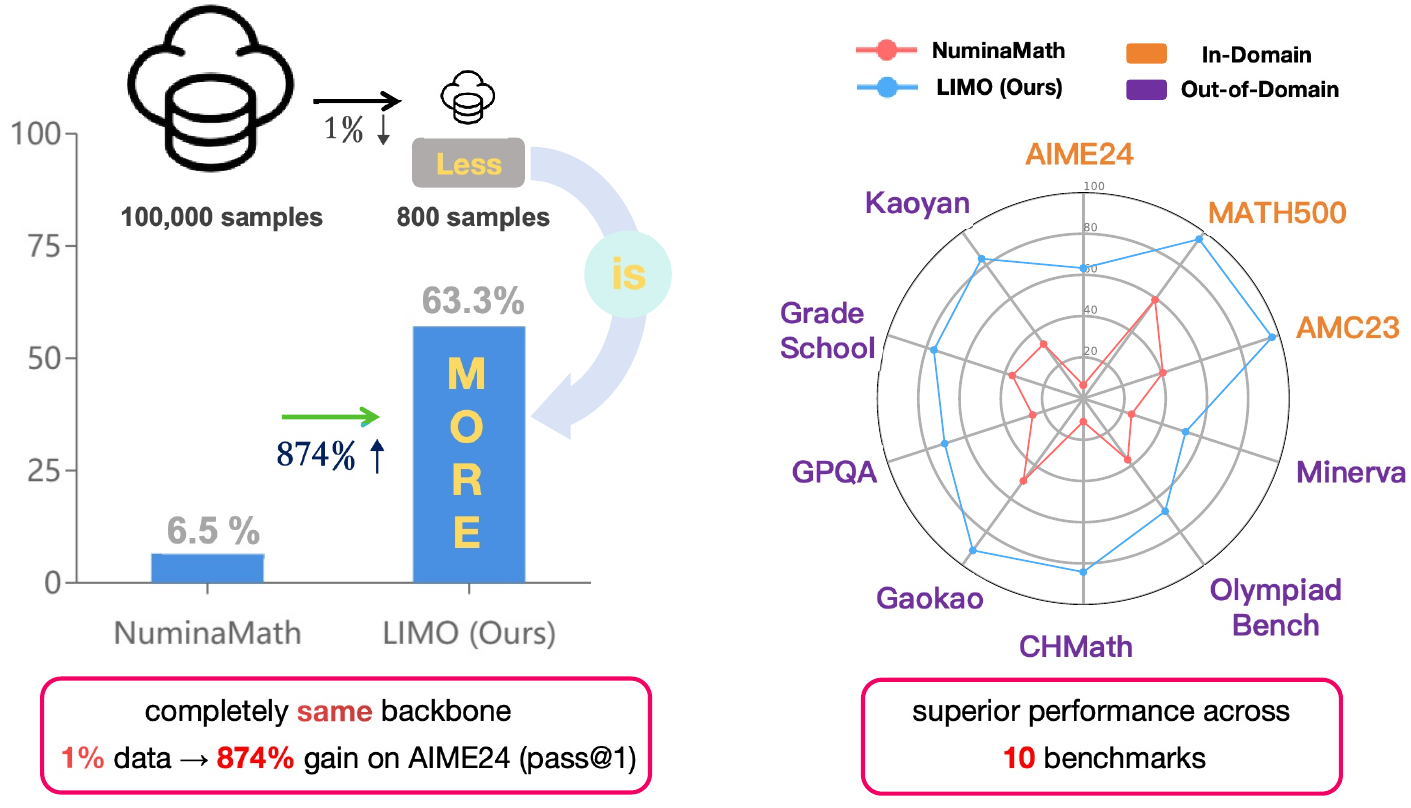}
        \caption{LIMO achieves substantial improvement over NuminaMath with fewer
        samples while excelling across diverse mathematical and multi-discipline
        benchmarks.}
        \label{fig:enter-label}
    \end{figure}

    \section{Introduction}

    Complex reasoning has long been considered one of the most challenging
    capabilities to instill in large language models (LLMs). While recent work has
    shown that LLMs can be effectively aligned with user preferences through relatively
    small amounts of instruction data~\citep{zhou2023lima}, teaching models to
    reason—particularly in mathematics and programming—is widely believed to
    require vastly more training examples~\citep{paster2023openwebmathopendatasethighquality,yue2024mammoth2scalinginstructionsweb}.
    This conventional wisdom stems from the inherent complexity of reasoning tasks,
    which demand multi-step logical deduction, domain knowledge application, and
    structured solution paths. The resulting paradigm typically involves training
    on tens or hundreds of thousands of examples~\citep{yu2024metamathbootstrapmathematicalquestions,li2024numinamath},
    based on two fundamental assumptions: first, that mastering such complex
    cognitive processes requires extensive supervised demonstrations, and second,
    that supervised fine-tuning leads primarily to memorization rather than true generalization~\citep{zhang2024carefulexaminationlargelanguage,xu2024benchmarkingbenchmarkleakagelarge,chu2025sftmemorizesrlgeneralizes}.

    While this approach has shown success, it imposes substantial computational costs. We argue this data-intensive paradigm may no longer be necessary. Recent advances have transformed how LLMs acquire and utilize reasoning knowledge, suggesting a more efficient approach. Two key developments have created conditions for reimagining reasoning in LLMs:
    \textbf{(1) Knowledge foundation revolution}: Modern foundation models now incorporate unprecedented amounts of mathematical content during pre-training~\citep{qwen2025qwen25technicalreport,yang2024qwen2,wang2024mathpilebilliontokenscalepretrainingcorpus}. Llama 2's total training data was 1.8T tokens~\citep{touvron2023llama2openfoundation}, while Llama 3 used 3.7T tokens for mathematical reasoning~\citep{grattafiori2024llama3herdmodels}. Contemporary LLMs may already possess rich mathematical knowledge, transforming the challenge from knowledge acquisition to knowledge elicitation.
    \textbf{(2) Inference-time computation scaling revolution}: Techniques scaling longer reasoning chains reveal that effective reasoning requires substantial computational space during inference. Recent works~\citep{openai2024openaio1card,qin2024o1,huang2024o1,guo2025deepseek} show that extended reasoning chains significantly improve reasoning ability. Inference-time computation provides the crucial \emph{cognitive workspace} where models can unpack and apply their pre-trained knowledge.

    We hypothesize that successful reasoning emerges from the synergy of rich pre-trained knowledge and sufficient computational resources at inference time. These developments suggest that if models possess rich reasoning knowledge and adequate computational space, activating their reasoning capabilities may require only a small number of high-quality samples that encourage extended deliberation, rather than massive fine-tuning datasets. We propose the \textbf{Less-Is-More Reasoning (LIMO) Hypothesis}, identifying two critical factors determining the \emph{elicitation threshold} for complex reasoning: (1) the latent presence of prerequisite knowledge within the model's parameters, and (2) the effectiveness of minimal exemplars in demonstrating problem-solving processes that encourage extended deliberation. The sample efficiency of eliciting advanced reasoning is thus bounded by the model's encoded knowledge foundation and its exposure to training samples that effectively utilize inference-time computation space.

    Our LIMO approach begins with a rigorous data curation process designed to identify high-quality samples that maximize reasoning elicitation. Starting from a large pool of QA pairs, we implement a multi-layered filtering system: first conducting coarse difficulty filtering to eliminate trivial problems, then fine-grained difficulty assessment to identify challenging questions, followed by knowledge point diversification to ensure comprehensive coverage. Simultaneously, we filter reasoning chains based on logical coherence, step-by-step clarity, and solution accuracy. This meticulous process yields a compact yet potent dataset of just \textbf{800} training samples. With simple supervised fine-tuning (SFT) using only this curated dataset based on Qwen2.5-32B-Instruct~\citep{qwen2025qwen25technicalreport}, LIMO achieves 63.3\% accuracy on the highly challenging AIME benchmark and 95.6\% on MATH, outperforming previous strong SFT-based models while using just 1\% of their training data. These benefits generalize across diverse previously unseen scenarios, with LIMO consistently outperforming models trained on 100x more data. This demonstrates that complex reasoning abilities can be effectively elicited through minimal but carefully curated training samples.

    The main contributions of this work are:
    (1) We establish the LIMO hypothesis, demonstrating that complex reasoning capabilities can be elicited with just hundreds of examples by leveraging rich mathematical knowledge in pre-trained models and detailed reasoning chains.
    (2) Following LIMO principles, we carefully construct the LIMO dataset and fine-tune Qwen2.5-32B-Instruct through simple SFT. Our experiments demonstrate that LIMO achieves highly-competitive performance on challenging mathematical reasoning benchmarks and maintains superior out-of-distribution performance.
    (3) Through extensive analysis and ablation studies, we validate the effectiveness of LIMO's data selection principles and explore their applicability across different scenarios (varying foundation model knowledge, model sizes, and architectures). Additionally, we investigate LIMO's minimum data requirements to achieve competitive performance.
    (4) We release our models, code, and curated datasets to support future research in data-efficient reasoning.

\section{Related Work}

    \subsection{Evolution of Mathematical Reasoning in LLMs}
    Large-scale training data has driven reasoning abilities in LLMs. During pretraining, reasoning is enhanced by relevant corpora~\citep{wang2024mathpilebilliontokenscalepretrainingcorpus,azerbayev2024llemmaopenlanguagemodel,paster2023openwebmathopendatasethighquality,shao2024deepseekmath} from textbooks, scientific papers, and mathematical code that capture diverse human cognitive patterns. Post-training research focuses on curating large-scale instruction data to teach reasoning~\citep{yue2023mammothbuildingmathgeneralist,yue2024mammoth2scalinginstructionsweb,li2024common7blanguagemodels,yu2024metamathbootstrapmathematicalquestions} by scaling questions and solutions. While this approach has achieved significant gains, it has been criticized for relying on memorization rather than true generalization~\citep{mirzadeh2024gsmsymbolicunderstandinglimitationsmathematical,zhang2024carefulexaminationlargelanguage}. \citet{mirzadeh2024gsmsymbolicunderstandinglimitationsmathematical} found that LLMs exhibit variance when responding to different instantiations of the same question, with performance declining when only numerical values are altered. This raises doubts about SFT methods' generalization capability~\citep{chu2025sftmemorizesrlgeneralizes} and whether LLMs can be true reasoners rather than merely knowledge retrievers~\citep{Kambhampati_2024}.

    \subsection{Test-time Scaling and Long Chain Reasoning}

    Instead of focusing on scaling model parameters and training data~\citep{kaplan2020scalinglawsneurallanguage},
    recent work has shifted to exploring test-time scaling~\citep{openai-o1,snell2024scaling},
    i.e., increasing the number of tokens to improve performance. This can be achieved
    by augmenting LLMs with methods such as parallel sampling~\citep{brown2024largelanguagemonkeysscaling,
    wang2022self, Li_2022} or symbolic tree search~\citep{hao2023reasoninglanguagemodelplanning,
    chen2024alphamath, yao2023treethoughtsdeliberateproblem} to enhance reasoning
    ability. Furthermore,~\citet{openai-o1, guo2025deepseek} explore training
    LLMs using reinforcement learning to generate long CoT, which often include self-reflection,
    verification, and backtracking—processes commonly employed by humans when solving
    complex problems. This approach not only innovates the training paradigm for
    LLMs but also provides a new form of training data to augment their
    reasoning ability. Our work demonstrates that this long CoT exhibits high-quality
    characteristics in eliciting the inherent reasoning abilities of LLMs.

    \subsection{Data Efficiency in Language Models}

    \citet{zhou2023lima} demonstrates that with just 1,000 carefully curated prompts
    and responses, models can learn to follow specific formats and generalize
    well to unseen tasks. The findings emphasize the importance of quality over quantity
    in the alignment process. However, whether this lesson can be applied to reasoning
    tasks remains uncertain, given the potential high computational complexity
    of such tasks~\citep{merrill2024expressivepowertransformerschain,xiang20252reasoningllmslearning}.
    While some work on reasoning highlights the importance of quality during the
    curation of training data~\citep{zhou2024programmingexampleliftingpretraining,yu2024metamathbootstrapmathematicalquestions},
    the quantity of such data is still much larger compared to that in LIMA. Our
    work extends the ideology of LIMA to reasoning tasks by investigating what constitutes
    high-quality questions and solutions, and demonstrates that the reasoning
    ability of LLMs can be enhanced in a highly data-efficient manner.

\section{LIMO Dataset}

    We formalize the \textbf{Less-Is-More Reasoning (LIMO) Hypothesis} as follows: In
    foundation models where domain knowledge has been comprehensively encoded
    during pre-training, sophisticated reasoning capabilities can emerge through
    minimal but precisely orchestrated demonstrations of cognitive processes.
    This hypothesis rests on two fundamental premises: (I) The latent presence of
    prerequisite knowledge within the model's parameter space (II) The quality of
    reasoning chains that precisely decompose complex problems into detailed,
    logical steps, making the cognitive process explicit and traceable. To
    validate this hypothesis, we propose a systematic approach to construct a high-quality,
    minimal dataset that can effectively elicit the model's inherent reasoning
    capabilities.

    \subsection{High-Quality Data Curation}
    In this paper, we focus on reasoning tasks with verifiable answers. Given a question $q \in \mathcal{Q}$, we aim to generate an answer $a \in \mathcal{A}$ via a reasoning chain $r \in \mathcal{R}$ consisting of intermediate steps $\{s_{1}, s_{2}, ..., s_{n}\}$. This reasoning process is formalized as $f: \mathcal{Q}\rightarrow \mathcal{R}\times \mathcal{A}$. Dataset quality depends on both question quality and solution quality. Our curation process constructs a deliberately small, high-quality dataset $\mathcal{D} = \{(q_{i}, r_{i}, a_{i})\}_{i=1}^{N}$ to validate our LIMO hypothesis that prioritizes quality over quantity.

    \begin{figure}[ht]
        \centering
        \includegraphics[width=1.0\linewidth]{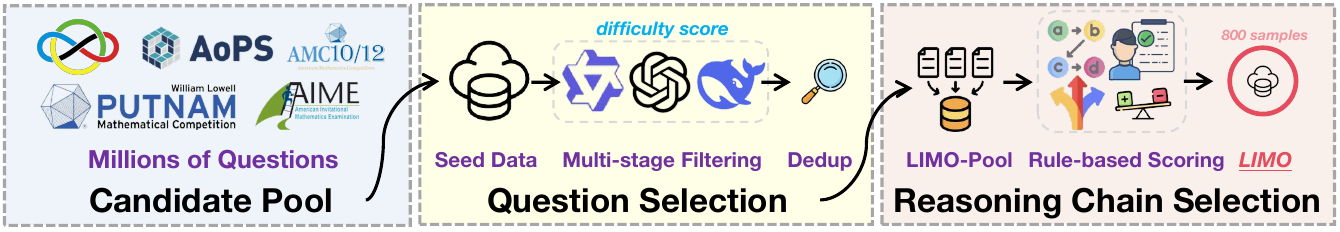}
        \caption{The LIMO dataset construction pipeline}
        \label{fig:data_pipeline}
    \end{figure}

    \subsubsection{Question Selection}
    \label{Question Selection}
    We hypothesize that high-quality questions $q \in \mathcal{Q}$ should naturally elicit extended reasoning processes. Our selection criteria prioritize challenging problems that foster complex reasoning chains and knowledge integration, while also considering problems that promote exploration of diverse problem-solving approaches. 
    
    To implement these criteria effectively, we first assembled a comprehensive pool of candidate problems from various established datasets: NuminaMath-CoT~\citep{li2024numinamath}, featuring meticulously annotated problems from high school to advanced competition levels; DeepScaleR~\citep{deepscaler2025}, consists of approximately 40,000 unique mathematics problem-answer pairs; AIME historical examination problems before 2024, known for its extremely challenging and integrative problems spanning multiple mathematical domains; MATH~\citep{hendrycks2021measuring}, encompassing various competitive mathematics problems from prestigious contests; and several questions including Chinese elementary, middle school, high school, and undergraduate-level exercises and examination papers.

    From this extensive initial collection, we implemented a systematic multi-stage filtration pipeline (Figure~\ref{fig:data_pipeline}). Starting with a corpus of tens of millions of mathematical problems, we first applied a baseline difficulty filter using a short-CoT mathematical model, Qwen2.5-Math-7B-Instruct~\citep{yang2024qwen2}. Problems that this model solved correctly within four attempts were excluded, ensuring that only non-trivial problems remained. Next, we subjected the filtered problems to a more rigorous evaluation using a stronger reasoning model, DeepSeek-R1-Distill-Qwen-32B~\citep{guo2025deepseek}. For each remaining problem, we sampled 32 solution attempts and used the empirical success rate as a difficulty indicator. Problems that were successfully solved in only 1–3 out of 32 attempts were retained, yielding a refined dataset of 2,125 problems, which composed our initial data pool (referred to as LIMO-Pool). To ensure dataset integrity, we conducted thorough deduplication against all evaluation benchmarks with n-gram matching, confirming no overlap existed.

    \subsubsection{Reasoning Chain Construction}
    \label{Reasoning Chain Construction}
    The quality of reasoning chains fundamentally impacts the effectiveness of large language model training. To develop our solution dataset, we employed three state-of-the-art reasoning models—DeepSeek R1, DeepSeek-R1-Distill-Qwen-32B~\citep{guo2025deepseek}, and QwQ-32B~\citep{qwq32b}—sampling multiple solutions from each to generate diverse reasoning approaches. Subsequently, all the authors conducted a comprehensive analysis of these filtered solutions through collaborative examination. Through careful observation and systematic review, we identified several key characteristics that distinguish high-quality reasoning chains:
    
    \textbf{Elaborated Reasoning}: Comprehensive exploration of logical steps without premature conclusions
    
    \textbf{Self-Verification}: Regular validation of intermediate results and logical consistency
    
    \textbf{Exploratory Approach}: Consideration of multiple possibilities before reaching conclusions
    
    \textbf{Adaptive Granularity}: Appropriate detail level across simple and complex deductions

    To quantify these qualities, we implemented a rule-based scoring system that calculated weighted metrics for each dimension. Elaborated Reasoning was measured by solution length (30\% weight); Self-Verification through frequency of validation-related words like "check" and "verify" (20\% weight); Exploratory Approach by counting tentative expressions such as "perhaps" and "might" (25\% weight); and Adaptive Granularity via connective phrases like "therefore" and "since" (25\% weight). All keyword frequencies were normalized by text length to ensure fair comparison across solutions of different sizes.
    
    From our initial collection of 2125 questions, we selected the highest-scoring solution for each problem, ranked these pairs by quality score, and extracted the top \textbf{800} to form the LIMO Dataset. This curation process embodies our \textbf{Less-Is-More} philosophy—prioritizing demonstration quality over quantity to enhance complex reasoning capabilities.

    \section{Training Recipe}
    Based on the \textbf{Less-Is-More} principle, a model with substantial reasoning knowledge from pre-training and the ability to perform long-chain reasoning at test time can develop robust reasoning abilities. With exposure to just a few hundred carefully selected SFT examples, the model learns to integrate meta-reasoning tasks into cohesive reasoning chains.
    
    We fine-tune Qwen2.5-32B-Instruct using supervised fine-tuning on our LIMO dataset. We set as our sequence length limit because all SFT response sequences remain under 16,384 tokens. The training process employs full-parameter fine-tuning with DeepSpeed ZeRO-3 optimization \citep{rajbhandari2020zero} and FlashAttention-2 \citep{dao2023flashattention}. For optimization, we utilize a learning rate of 5.0e-6 with a cosine decay schedule. We deliberately omit the warmup phase to facilitate rapid adaptation to the high-quality reasoning examples in our dataset. The model is trained for 15 epochs with a batch size of 64 examples to balance computational efficiency and stable convergence.
    
    \section{Evaluation Framework}

    We establish a comprehensive evaluation framework to assess our models' mathematical reasoning capabilities across various dimensions. Our framework encompasses both in-domain and out-of-distribution evaluations, utilizing established benchmarks alongside novel multilingual tests to thoroughly examine generalization capabilities beyond the training distribution.
    
    \paragraph{In-domain Evaluation}
    
    To comprehensively assess the models' performance across various reasoning capabilities, we have established a diverse evaluation framework encompassing both traditional and novel benchmarks. Our primary evaluation suite includes several well-established mathematical competitions and benchmarks: the American Invitational Mathematics Examination (AIME24), MATH500~\citep{hendrycks2021measuring}, and the American Mathematics Competitions (AMC23).
    
    \paragraph{Out-of-distribution Evaluation}
To rigorously evaluate OOD performance, we select benchmarks differing from our training data across three categories. First, we include diverse mathematical competitions like OlympiadBench~\citep{he2024olympiadbench} to test performance on different mathematical challenges. Second, to minimize data contamination, we construct novel multilingual benchmarks using recent examination problems: CHMath (2024 Chinese High School Mathematics League), Gaokao (China's 2024 College Entrance Exam), Kaoyan (Chinese Graduate School Entrance Exams), and GradeSchool (our new elementary mathematics benchmark). These Chinese-language problems introduce an additional OOD dimension, assessing both cross-distribution and cross-lingual reasoning capabilities. Third, we incorporate multi-disciplinary benchmarks including Minerva~\citep{lewkowycz2022solving} (undergraduate-level STEM) and GPQA~\citep{rein2023gpqa} to evaluate the transfer of mathematical reasoning skills to broader contexts beyond our training domain.
    
    \paragraph{Performance metrics}
    
    We evaluate performance using the pass@1 metric in a zero-shot chain-of-thought setting across all benchmarks. For larger benchmarks (MATH500, OlympiadBench, Gaokao, Kaoyan, GradeSchool, MinervaMath, and GPQA), we employ greedy decoding with a single sample. For smaller benchmarks (less than 50 problems: AIME24, AMC23, and CHMath), we generate 4 samples (temperature=0.6) and calculate the unbiased pass@1 metric~\citep{chen2021evaluating}. We use rule-based evaluations for numerical answers and an LLM-based evaluator for complex answer formats. All evaluations maintain a 32,768 token maximum output length.

    \section{Experiment}

    \subsection{Baselines}
    We compare LIMO against a comprehensive set of baselines with several prominent models. These include \textbf{OpenAI-o1-preview}~\citep{openai-o1}, a large language model that has demonstrated advanced mathematical reasoning abilities across various complex tasks; \textbf{QwQ-32B-Preview}~\citep{qwq}, a model specifically designed for mathematical problem-solving with strong reasoning capabilities; and \textbf{Qwen2.5-32B-Instruct}, which serves as our base model for comparative analysis.
    
    To investigate the impact of training data efficiency, we conduct comparative experiments using mainstream open-source reasoning datasets for supervised fine-tuning on our base model. For a fair comparison, all experiments use the same LLM backbone as LIMO, ensuring that performance differences are solely attributable to the training data characteristics. These comparative datasets include OpenThoughts-114k~\citep{openthoughts}, a synthetic reasoning dataset containing 114k examples covering mathematics, science, coding, and puzzles, with solutions following a structured reasoning format generated by DeepSeek-R1; and \textbf{NuminaMath-100k}, a randomly selected 100k subset of NuminaMath-CoT, featuring mathematical problems ranging from Chinese high school exercises to international mathematics olympiad competitions, with each solution following a Chain of Thought (CoT) format \citep{wei2022chain}.
    
    These datasets contain substantially more samples than LIMO's training set (800 examples), allowing us to examine the relationship between data quantity and model performance.

    \subsection{Main Results}

    \input{table/main-results}

    Our experimental results demonstrate LIMO's superior performance across both
    in-domain and out-of-domain tasks, as shown in Table~\ref{tab:main_results}.
    \paragraph{In-domain Performance}
    On in-domain tasks, LIMO achieves the best results across all benchmarks. For
    AIME24, LIMO achieves 63.3\% accuracy, outperforming QwQ-32B-Preview (50.0\%)
    and OpenAI-o1-preview (44.6\%) by significant margins. On
    MATH500, LIMO achieves a notable 95.6\% accuracy, surpassing QwQ-32B-Preview (89.8\%)
    and OpenAI-o1-preview (85.5\%). The performance gap is even more pronounced
    on AMC23, where LIMO reaches 96.3\% accuracy compared to QwQ-32B-Preview's
    83.6\%.
    \paragraph{Out-of-domain Generalization}
    LIMO demonstrates strong generalization capabilities across diverse out-of-domain
    tasks. On OlympiadBench, LIMO achieves 67.6\% accuracy, significantly outperforming
    QwQ-32B-Preview (58.5\%) and the base model (45.3\%). Similar improvements
    are observed on other challenging benchmarks such as CHMath (84.2\% vs 68.5\%)
    and GradeSchool (76.2\% vs 63.8\%). Notably, LIMO maintains competitive
    performance even on GPQA, where it achieves 70.7\% accuracy, close to OpenAI-o1-preview's
    leading score of 73.3\%.
    \paragraph{Comparison with Larger Datasets}
    Our experiments reveal that despite larger scale, both baseline datasets underperform
    compared to LIMO. NuminaMath-100k shows significant degradation (32.3\% vs. base
    model's 49.9\%) due to uncurated reasoning chains, while OpenThoughts-114k
    achieves suboptimal results (58.3\%) probably due to unfocused problem selection.
    In contrast, LIMO's carefully curated 800 problems yield superior performance
    (78.1\%), demonstrating that targeted selection are more crucial than data quantity for developing robust reasoning capabilities.

    \paragraph{Overall Performance}
    LIMO achieves the highest average performance of 78.1\% across all benchmarks,
    substantially outperforming OpenAI-o1-preview, QwQ-32B-Preview,
    and other baselines. This comprehensive evaluation demonstrates that LIMO's carefully
    curated training approach with just 800 examples can outperform models trained
    on datasets that are orders of magnitude larger.

    \subsection{Analysis}

 \subsubsection{RQ1: Impact of Reasoning Chain Quality}

To further validate and compare the effectiveness of our LIMO method, we first focus on the selection of high-quality reasoning chains. We investigate what characteristics define superior reasoning chains that lead to better model performance through a controlled comparative study of solutions with varying quality for identical problems. \input{table/cot_quality}
For this analysis, we select 500 questions from the LIMO dataset that each have multiple correct solutions generated by diverse models, ensuring we have varied yet accurate approaches to the same problems. 
We collect and categorize these solutions into five distinct quality levels (L1-L5, with L5 being the highest) based on our rule-based scoring system described in Section~\ref{Reasoning Chain Construction}.

Results in Figure \ref{fig:data_quality} show clear correlation between reasoning quality and model performance. L5-trained models achieve highest results on both AIME24 and MATH500, with performance decreasing consistently with each quality level. The substantial gap between L5 and L1 solutions demonstrates that reasoning chain quality significantly influences model performance, underscoring the importance of curating high-quality training data.

\subsubsection{RQ2: Impact of Question Quality}
\begin{figure}[ht]
    \centering
    \includegraphics[width=0.8\linewidth]{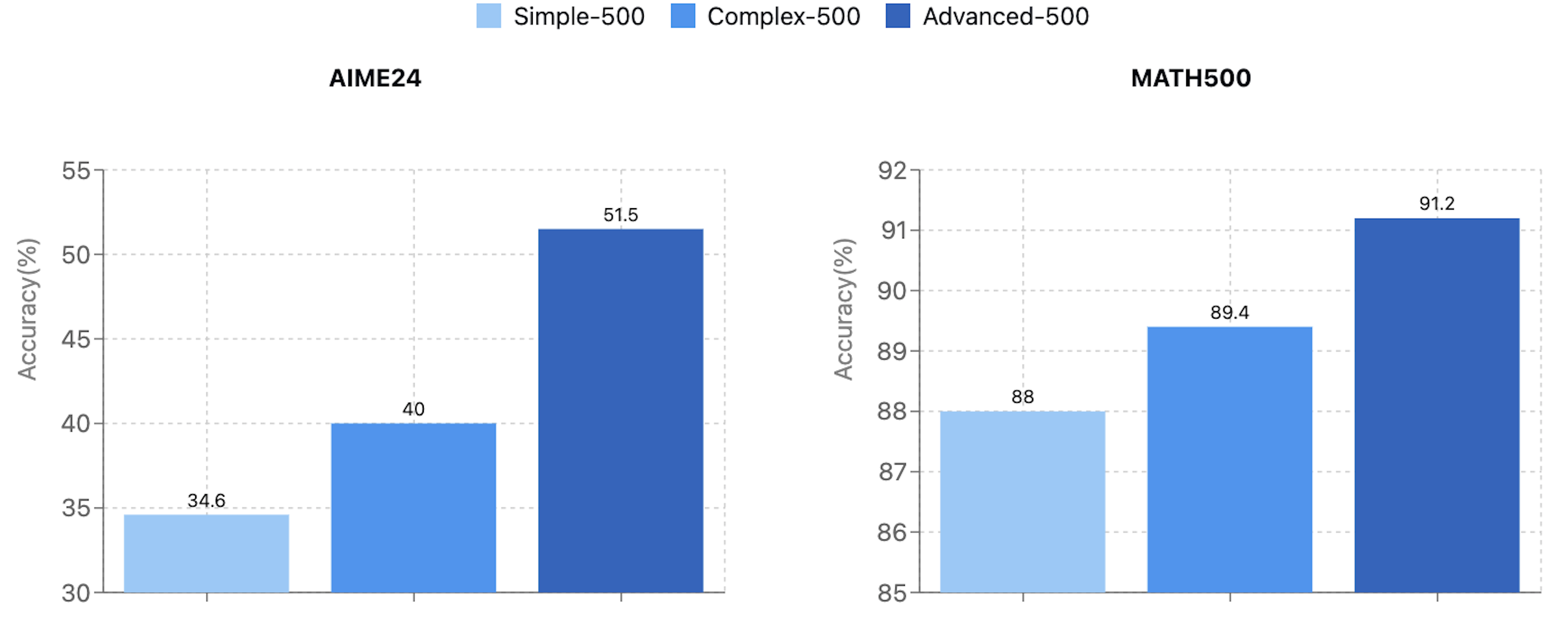}
    \caption{Performance comparison on MATH and AIME benchmarks between models trained on different question quality: Simple-500, Complex-500, and Advanced-500.}
    \label{fig:question_quality}
\end{figure}

We hypothesize that more challenging problems foster complex reasoning chains and enhanced knowledge integration. To test this, we examine how question difficulty affects models' reasoning capabilities. We select three sets of 500 problems with increasing difficulty: \textbf{Simple-500} (MATH levels 1-2), \textbf{Complex-500} (MATH levels 3-5), and \textbf{Advanced-500} (AIME problems). We verify the difficulty gradient by evaluating various LLMs, observing declining accuracy and increasing solution length across these sets. We then use DeepSeek-R1 to generate high-quality solutions for each set and fine-tune Qwen2.5-32B-Instruct on them. Results in Figure \ref{fig:question_quality} show that modifying problem selection alone leads to 16\% accuracy improvement on AIME2024, reaching 51.5\%. Notably, the model fine-tuned on Advanced-500 achieves 91.2\% on MATH500 despite no in-domain training data, suggesting that reasoning improvements from increased problem difficulty generalize across datasets.

\subsubsection{RQ3: LLM Backbone - Pre-trained Knowledge}

Building on our LIMO hypothesis that emphasizes latent prerequisite knowledge within model parameters, we examine how pre-training data affects a model's ability to leverage minimal exemplars for math reasoning. To isolate pre-training impact while controlling for architecture and fine-tuning, we compare two 32B-parameter models: Qwen1.5-32B-Chat \citep{qwen1.5} and Qwen2.5-32B-Instruct (LIMO's base model). While sharing identical architecture, Qwen2.5 features improved pre-training data quality, particularly in mathematical and code-related content. We SFT both models using identical LIMO datasets and evaluate on AIME2024 and MATH500 benchmarks.

\input{table/backbone}

Results in Figure~\ref{fig:llm_backbone} show pre-trained model choice dramatically impacts reasoning performance. LIMO (built on Qwen2.5) achieves 63.3\% accuracy on AIME2024, a 54.1 percentage point improvement over Qwen1.5's 9.2\%. On MATH500, LIMO reaches 95.6\% accuracy, surpassing Qwen1.5 by 30.4 percentage points. These substantial improvements confirm that Qwen2.5's enhanced pre-training creates a stronger foundation for mathematical reasoning, aligning with our hypothesis that richer pre-trained knowledge enables more effective utilization of minimal exemplars.

It is hypothesized that increasing the number of parameters in a large language model enhances its capacity for deep reasoning and overall performance in complex tasks. To investigate this hypothesis, we fine-tune models from the Qwen2.5-Instruct series of varying sizes—3B, 7B, 14B, 32B, and 72B—using the LIMO dataset with 800 high-quality samples. The models are subjected to the same supervised fine-tuning  recipe to ensure that any observed differences in performance are primarily attributable to model size.

\subsubsection{RQ4: LLM Backbone - Model Size}    

\begin{figure}[H]
    \centering
    \includegraphics[width=1\linewidth]{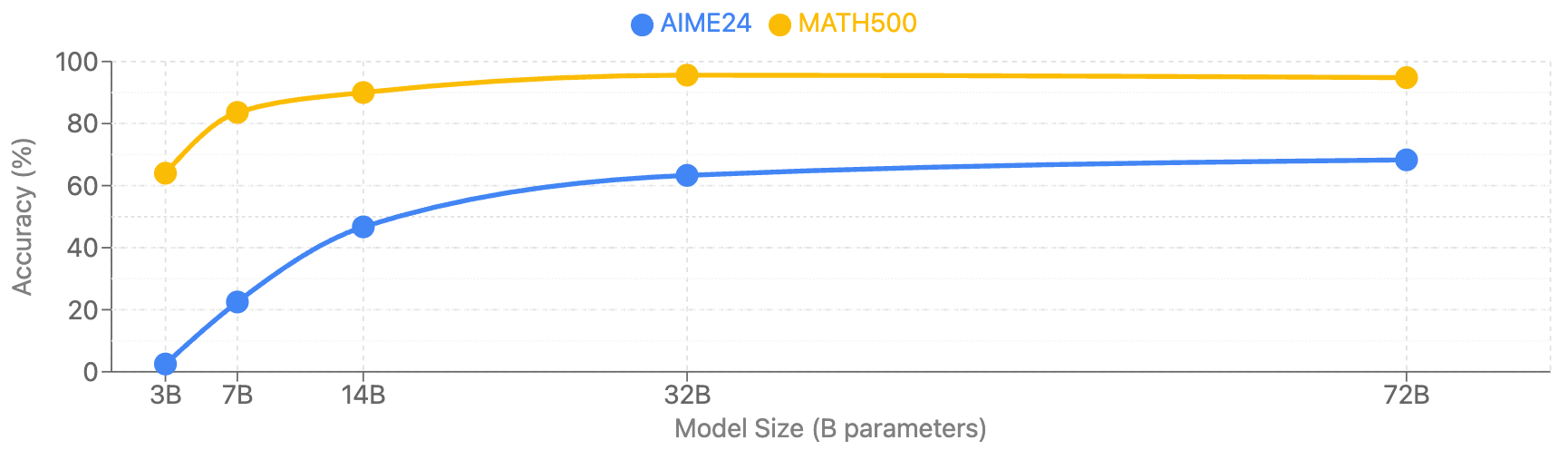}
    \caption{Scaling of mathematical reasoning ability with  model size}
    \label{fig:model_size}
\end{figure}

Figure~\ref{fig:model_size} summarizes the performance of the models on both benchmarks. The performance on AIME24 shows a marked increase with model size, rising from 2.5 for the 3B model to 68.3 for the 72B model. This supports the hypothesis that larger models are better able to handle the deep reasoning required by competition-level math problems.
While the improvements on MATH500 are also evident, the gains are less dramatic, suggesting that even smaller models can achieve high accuracy on easier benchmarks like MATH500.
The difference in performance between the 32B and 72B models is marginal on MATH500 (95.6\% vs. 94.8\%) and slightly less pronounced on AIME24 (63.3\% vs. 68.3\%). This indicates a potential saturation point, where increasing the number of parameters further may yield diminishing returns in fine-tuning performance for these benchmarks.

\subsubsection{RQ5: Sample Efficiency}
Our experiments reveal that a surprisingly small number (i.e. 800) of samples can elicit competition-level mathematical reasoning, though the lower bound for maintaining effective performance remains an open question. We explore the impact of dataset size on fine-tuning efficacy by systematically varying the number of training samples. To optimize performance with constrained dataset sizes, we rank all 2,125 questions in the LIMO-pool dataset based on their best reasoning chains' quality scores. From this ranked set, we select top-ranked subsets of varying sizes—400, 800, 1,200, 1,600, and 2,000 questions—to construct five datasets (LIMO-400 through LIMO-2k). Each dataset is used to fine-tune our base model following an identical training recipe to ensure comparability.

\begin{wrapfigure}[19]{r}{0.4\textwidth}
\includegraphics[width=\linewidth]{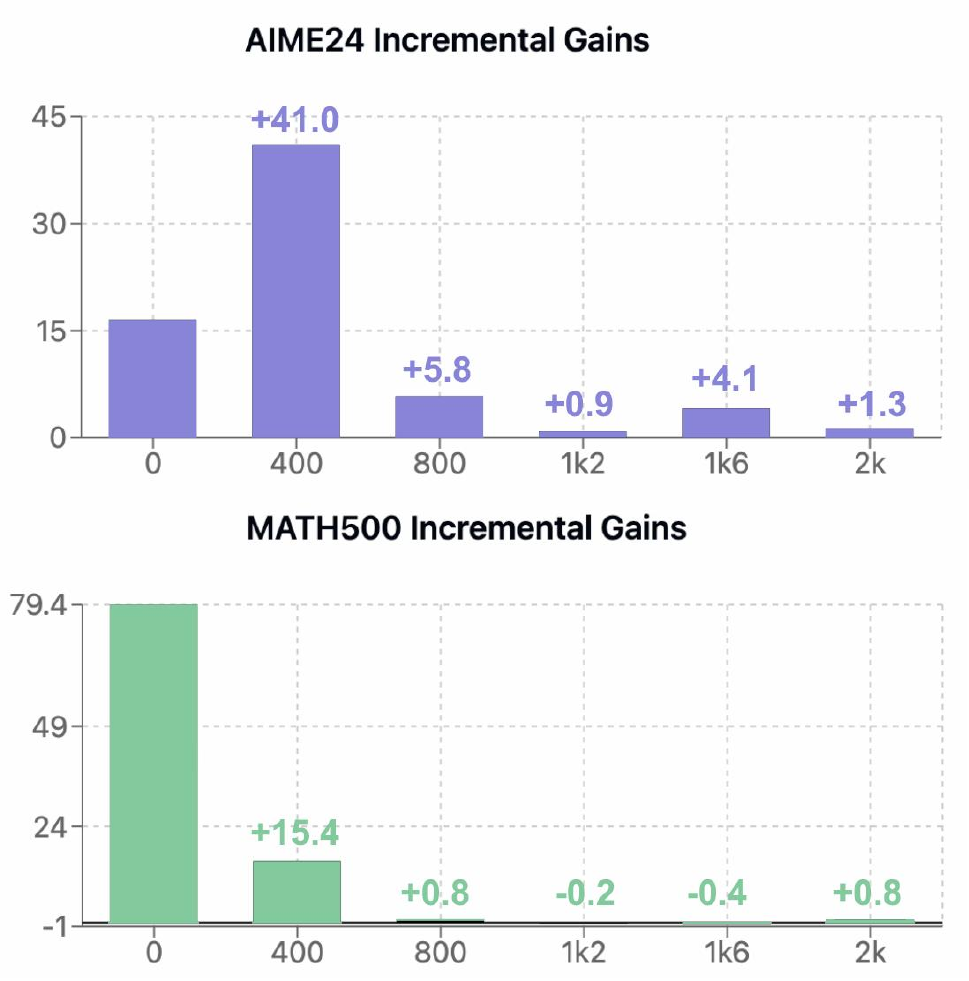}
\caption{Incremental performance gains by dataset size.}
\label{fig:sample-efficiency-gain}
\end{wrapfigure}

\begin{figure}[h!]
    \centering
    \includegraphics[width=1\linewidth]{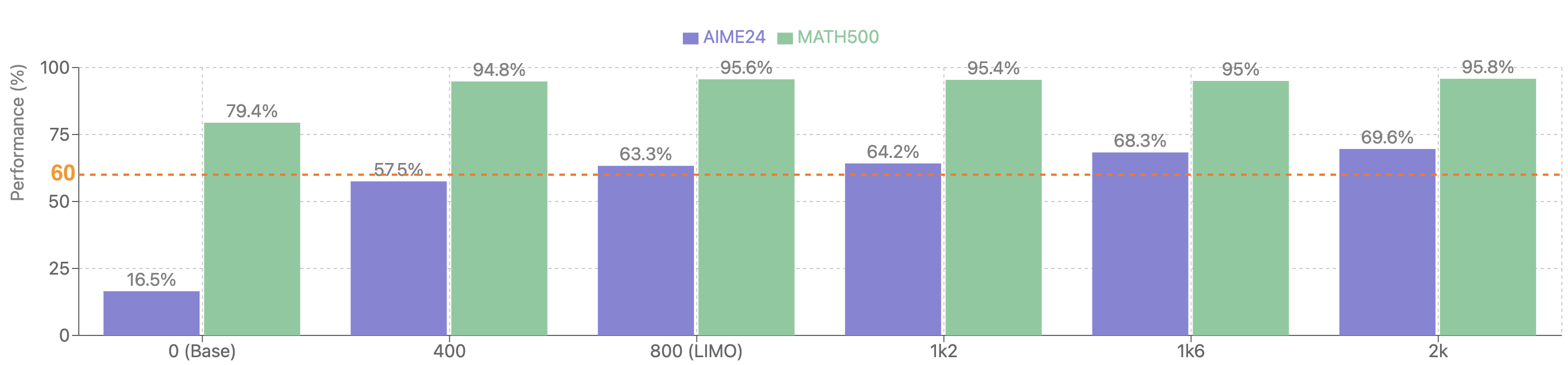}
    \caption{Impact of dataset size on model performance}
    \label{fig:sample-efficiency}
\end{figure}

Figure \ref{fig:sample-efficiency-gain} and Figure \ref{fig:sample-efficiency} present the performance of models fine-tuned on different dataset sizes. Several key observations emerge: (1) Fine-tuning with just 400 samples yields dramatic improvement over the base model, increasing AIME24 accuracy from 16.5 to 57.5 and MATH500 accuracy from 79.4 to 94.8. (2) While performance continues to improve with larger datasets, we observe diminishing returns beyond 800 samples, with only marginal increases between LIMO-800 and LIMO-1k2 (+0.9 on AIME24, -0.2 on MATH500), suggesting an early plateau effect. (3) The highest dataset size (2k) provides the best overall performance (69.6 on AIME24, 95.8 on MATH500), but with minimal improvement over 1.6k, indicating that beyond a certain threshold, additional data contributes less significantly to fine-tuning gains.
These findings highlight the efficiency of high-quality, carefully selected data in enhancing LLM mathematical reasoning. Future work could explore active learning strategies to further optimize sample efficiency.

\section{Conclusion}
Our work demonstrates that complex mathematical reasoning in LLMs can be achieved with surprisingly few examples, contradicting the prevailing assumption that massive training data is necessary. LIMO achieves competitive performance on challenging benchmarks using only 1\% of the training data required by previous approaches. This confirms our Less-Is-More Reasoning Hypothesis: in knowledge-rich foundation models, sophisticated reasoning emerges through minimal but precisely orchestrated demonstrations that effectively utilize inference-time computation.

\section{Acknowledgement}
We would like to express our sincere gratitude to Yixiu Liu and Yiwei Qin for their valuable contributions to this research work. Their expertise, dedication, and collaborative spirit have significantly enhanced the quality of our study. Their insightful suggestions and technical assistance were instrumental in achieving our research objectives. We are also grateful to Run-Ze Fan for his diligent efforts in labeling our internal benchmark.
We also wish to extend our appreciation to Haoyang Zou and Xuefeng Li for their valuable discussions during the early stages of this work. Their perspectives and insights helped shape the foundation of our research.

\newpage

\bibliography{colm2025_conference}
\bibliographystyle{colm2025_conference}

\end{document}

%% file: table/main-results.tex
\begin{table*}[ht]
    \scalebox{0.8}{
    \begin{tabular}{l c c c c c c}
        \toprule
        \textbf{Datasets} & \textbf{\makecell{OpenAI-o1\\-preview}} & \textbf{\makecell{Qwen2.5-32B\\-Instruct}} & \textbf{\makecell{QwQ-32B-\\preview}}  & 
        \textbf{\makecell{OpenThoughts\\(114k)}} & \textbf{\makecell{NuminaMath\\(100k)}} & \cellcolor[HTML]{dceffe}
        \textbf{\makecell{LIMO\\ours(800)}} \\
        \midrule
        \multicolumn{7}{c}{In Domain} \\
        \midrule
        AIME24 & 44.6 & 16.5 & 50.0 & 50.2 & 6.5 & \cellcolor[HTML]{dceffe} \textbf{63.3}\\
        MATH500 & 85.5 & 79.4 & 89.8 & 80.6 & 59.2 & \cellcolor[HTML]{dceffe} \textbf{95.6}\\
        AMC23 & 81.8 & 64.0 & 83.6 & 80.5 & 40.6 & \cellcolor[HTML]{dceffe} \textbf{96.3}\\
        
        \midrule
        \multicolumn{7}{c}{Out of Domain} \\
        \midrule
        OlympiadBench & 52.1 & 45.3 & 58.5 & 56.3 & 36.7 & \cellcolor[HTML]{dceffe} \textbf{67.6}\\
        CHMath & 50.0 & 27.3 & 68.5 & 74.1 & 11.2 & \cellcolor[HTML]{dceffe} \textbf{84.2}\\
        Gaokao & 62.1 & 72.1 & 80.1 & 63.2 & 49.4 & \cellcolor[HTML]{dceffe} \textbf{91.1} \\
        Kaoyan & 51.5 & 48.2 & 70.3 & 54.7 & 32.7 & \cellcolor[HTML]{dceffe} \textbf{83.9}\\
        GradeSchool & 62.8 & 56.7 & 63.8 & 39.0 & 36.2 & \cellcolor[HTML]{dceffe} \textbf{76.2} \\
        Minerva & 47.1 & 41.2 & 39.0 & 41.1 & 24.6 & \cellcolor[HTML]{dceffe} \textbf{52.2}\\
        GPQA & \textbf{73.3} & 48.0 & 65.1 & 42.9 & 25.8 & \cellcolor[HTML]{dceffe} 70.7\\
        \midrule
        AVG. & 61.1 & 49.9 & 66.9 & 58.3 & 32.3 & \cellcolor[HTML]{dceffe} \textbf{78.1}\\
        \bottomrule
    \end{tabular}
    }
    \caption{
    \textbf{Comparison of model performance (pass@1) across various mathematical reasoning benchmarks} Models include state-of-the-art LLMs (OpenAI-o1-preview, QwQ-32B-Preview), our base model (Qwen2.5-32B-Instruct), and models fine-tuned on different datasets. Training data sizes are shown in parentheses. 
    Best results for each benchmark are shown in bold. Our proposed LIMO model (highlighted in blue) achieves superior performance despite using significantly fewer training examples (800) compared to other fine-tuned models (more than 100k).
    }
    \label{tab:main_results}
\end{table*}

%% file: table/cot_quality.tex
\begin{wrapfigure}[16]{r}{0.4\textwidth}
\includegraphics[width=\linewidth]{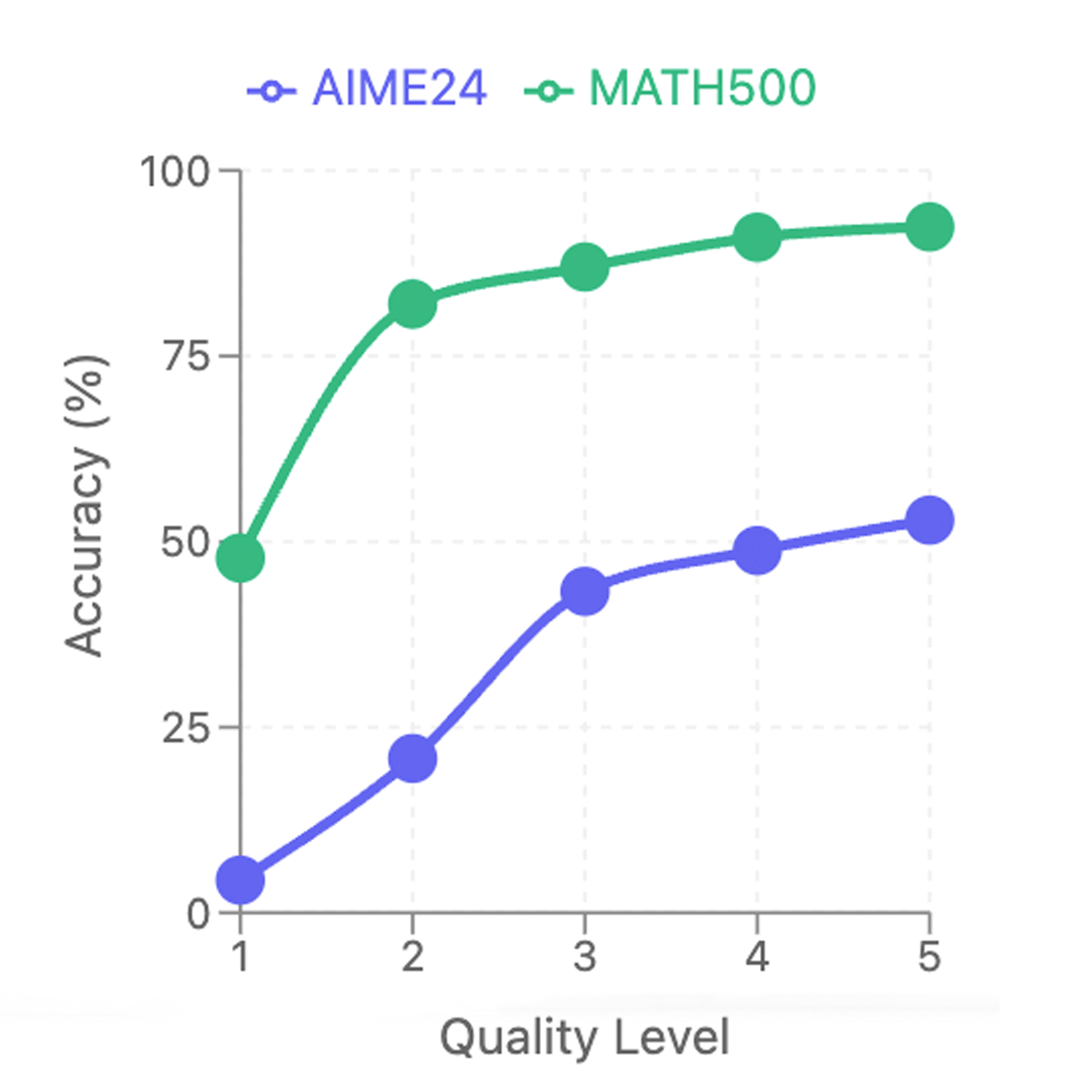}
\caption{Comparison of models trained on reasoning chains of different quality levels.}
\label{fig:data_quality}
\end{wrapfigure}

%% file: table/backbone.tex
\begin{wrapfigure}[19]{r}{0.275\textwidth}
\includegraphics[width=\linewidth]{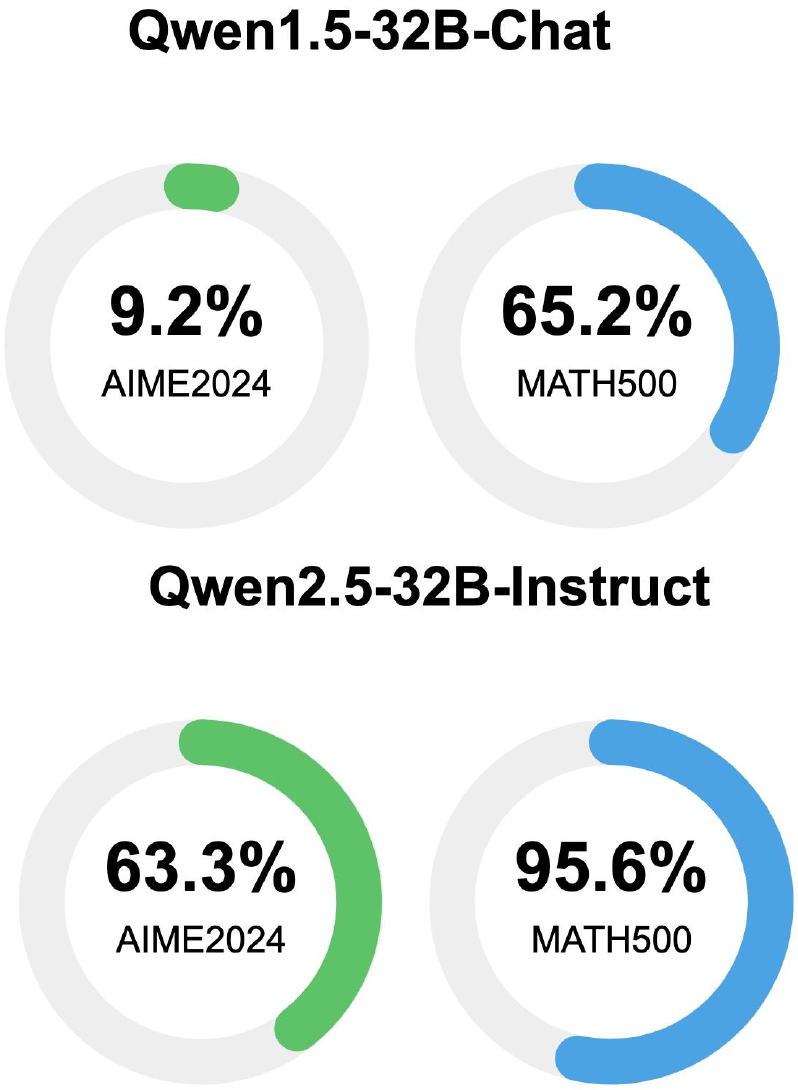}
\caption{Impact of pre-trained model choice on mathematical reasoning performance.}
\label{fig:llm_backbone}
\end{wrapfigure}

%% file: colm2025_conference.bbl
\begin{thebibliography}{48}
\providecommand{\natexlab}[1]{#1}
\providecommand{\url}[1]{\texttt{#1}}
\expandafter\ifx\csname urlstyle\endcsname\relax
  \providecommand{\doi}[1]{doi: #1}\else
  \providecommand{\doi}{doi: \begingroup \urlstyle{rm}\Url}\fi

\bibitem[Azerbayev et~al.(2024)Azerbayev, Schoelkopf, Paster, Santos, McAleer, Jiang, Deng, Biderman, and Welleck]{azerbayev2024llemmaopenlanguagemodel}
Zhangir Azerbayev, Hailey Schoelkopf, Keiran Paster, Marco~Dos Santos, Stephen McAleer, Albert~Q. Jiang, Jia Deng, Stella Biderman, and Sean Welleck.
\newblock Llemma: An open language model for mathematics, 2024.
\newblock URL \url{https://arxiv.org/abs/2310.10631}.

\bibitem[Brown et~al.(2024)Brown, Juravsky, Ehrlich, Clark, Le, Ré, and Mirhoseini]{brown2024largelanguagemonkeysscaling}
Bradley Brown, Jordan Juravsky, Ryan Ehrlich, Ronald Clark, Quoc~V. Le, Christopher Ré, and Azalia Mirhoseini.
\newblock Large language monkeys: Scaling inference compute with repeated sampling, 2024.
\newblock URL \url{https://arxiv.org/abs/2407.21787}.

\bibitem[Chen et~al.(2024)Chen, Liao, Li, and Fan]{chen2024alphamath}
Guoxin Chen, Minpeng Liao, Chengxi Li, and Kai Fan.
\newblock Alphamath almost zero: process supervision without process.
\newblock \emph{ArXiv preprint}, abs/2405.03553, 2024.
\newblock URL \url{https://arxiv.org/abs/2405.03553}.

\bibitem[Chen et~al.(2021)Chen, Tworek, Jun, Yuan, Ponde, Kaplan, Edwards, Burda, Joseph, Brockman, et~al.]{chen2021evaluating}
Mark Chen, Jerry Tworek, Heewoo Jun, Qiming Yuan, Henrique Ponde, Jared Kaplan, Harri Edwards, Yura Burda, Nicholas Joseph, Greg Brockman, et~al.
\newblock Evaluating large language models trained on code.
\newblock \emph{arXiv preprint arXiv:2107.03374}, 2021.

\bibitem[Chu et~al.(2025)Chu, Zhai, Yang, Tong, Xie, Schuurmans, Le, Levine, and Ma]{chu2025sftmemorizesrlgeneralizes}
Tianzhe Chu, Yuexiang Zhai, Jihan Yang, Shengbang Tong, Saining Xie, Dale Schuurmans, Quoc~V. Le, Sergey Levine, and Yi~Ma.
\newblock Sft memorizes, rl generalizes: A comparative study of foundation model post-training, 2025.
\newblock URL \url{https://arxiv.org/abs/2501.17161}.

\bibitem[Dao(2023)]{dao2023flashattention}
Tri Dao.
\newblock Flashattention-2: Faster attention with better parallelism and work partitioning.
\newblock \emph{arXiv preprint arXiv:2307.08691}, 2023.

\bibitem[Grattafiori et~al.(2024)Grattafiori, Dubey, Jauhri, Pandey, Kadian, Al-Dahle, Letman, Mathur, Schelten, Vaughan, Yang, Fan, Goyal, Hartshorn, Yang, Mitra, Sravankumar, Korenev, Hinsvark, Rao, Zhang, Rodriguez, Gregerson, Spataru, Roziere, Biron, Tang, Chern, Caucheteux, Nayak, Bi, Marra, McConnell, Keller, Touret, Wu, Wong, Ferrer, Nikolaidis, Allonsius, Song, Pintz, Livshits, Wyatt, Esiobu, Choudhary, Mahajan, Garcia-Olano, Perino, Hupkes, Lakomkin, AlBadawy, Lobanova, Dinan, Smith, Radenovic, Guzmán, Zhang, Synnaeve, Lee, Anderson, Thattai, Nail, Mialon, Pang, Cucurell, Nguyen, Korevaar, Xu, Touvron, Zarov, Ibarra, Kloumann, Misra, Evtimov, Zhang, Copet, Lee, Geffert, Vranes, Park, Mahadeokar, Shah, van~der Linde, Billock, Hong, Lee, Fu, Chi, Huang, Liu, Wang, Yu, Bitton, Spisak, Park, Rocca, Johnstun, Saxe, Jia, Alwala, Prasad, Upasani, Plawiak, Li, Heafield, Stone, El-Arini, Iyer, Malik, Chiu, Bhalla, Lakhotia, Rantala-Yeary, van~der Maaten, Chen, Tan, Jenkins, Martin, Madaan, Malo, Blecher,
  Landzaat, de~Oliveira, Muzzi, Pasupuleti, Singh, Paluri, Kardas, Tsimpoukelli, Oldham, Rita, Pavlova, Kambadur, Lewis, Si, Singh, Hassan, Goyal, Torabi, Bashlykov, Bogoychev, Chatterji, Zhang, Duchenne, Çelebi, Alrassy, Zhang, Li, Vasic, Weng, Bhargava, Dubal, Krishnan, Koura, Xu, He, Dong, Srinivasan, Ganapathy, Calderer, Cabral, Stojnic, Raileanu, Maheswari, Girdhar, Patel, Sauvestre, Polidoro, Sumbaly, Taylor, Silva, Hou, Wang, Hosseini, Chennabasappa, Singh, Bell, Kim, Edunov, Nie, Narang, Raparthy, Shen, Wan, Bhosale, Zhang, Vandenhende, Batra, Whitman, Sootla, Collot, Gururangan, Borodinsky, Herman, Fowler, Sheasha, Georgiou, Scialom, Speckbacher, Mihaylov, Xiao, Karn, Goswami, Gupta, Ramanathan, Kerkez, Gonguet, Do, Vogeti, Albiero, Petrovic, Chu, Xiong, Fu, Meers, Martinet, Wang, Wang, Tan, Xia, Xie, Jia, Wang, Goldschlag, Gaur, Babaei, Wen, Song, Zhang, Li, Mao, Coudert, Yan, Chen, Papakipos, Singh, Srivastava, Jain, Kelsey, Shajnfeld, Gangidi, Victoria, Goldstand, Menon, Sharma, Boesenberg,
  Baevski, Feinstein, Kallet, Sangani, Teo, Yunus, Lupu, Alvarado, Caples, Gu, Ho, Poulton, Ryan, Ramchandani, Dong, Franco, Goyal, Saraf, Chowdhury, Gabriel, Bharambe, Eisenman, Yazdan, James, Maurer, Leonhardi, Huang, Loyd, Paola, Paranjape, Liu, Wu, Ni, Hancock, Wasti, Spence, Stojkovic, Gamido, Montalvo, Parker, Burton, Mejia, Liu, Wang, Kim, Zhou, Hu, Chu, Cai, Tindal, Feichtenhofer, Gao, Civin, Beaty, Kreymer, Li, Adkins, Xu, Testuggine, David, Parikh, Liskovich, Foss, Wang, Le, Holland, Dowling, Jamil, Montgomery, Presani, Hahn, Wood, Le, Brinkman, Arcaute, Dunbar, Smothers, Sun, Kreuk, Tian, Kokkinos, Ozgenel, Caggioni, Kanayet, Seide, Florez, Schwarz, Badeer, Swee, Halpern, Herman, Sizov, Guangyi, Zhang, Lakshminarayanan, Inan, Shojanazeri, Zou, Wang, Zha, Habeeb, Rudolph, Suk, Aspegren, Goldman, Zhan, Damlaj, Molybog, Tufanov, Leontiadis, Veliche, Gat, Weissman, Geboski, Kohli, Lam, Asher, Gaya, Marcus, Tang, Chan, Zhen, Reizenstein, Teboul, Zhong, Jin, Yang, Cummings, Carvill, Shepard, McPhie,
  Torres, Ginsburg, Wang, Wu, U, Saxena, Khandelwal, Zand, Matosich, Veeraraghavan, Michelena, Li, Jagadeesh, Huang, Chawla, Huang, Chen, Garg, A, Silva, Bell, Zhang, Guo, Yu, Moshkovich, Wehrstedt, Khabsa, Avalani, Bhatt, Mankus, Hasson, Lennie, Reso, Groshev, Naumov, Lathi, Keneally, Liu, Seltzer, Valko, Restrepo, Patel, Vyatskov, Samvelyan, Clark, Macey, Wang, Hermoso, Metanat, Rastegari, Bansal, Santhanam, Parks, White, Bawa, Singhal, Egebo, Usunier, Mehta, Laptev, Dong, Cheng, Chernoguz, Hart, Salpekar, Kalinli, Kent, Parekh, Saab, Balaji, Rittner, Bontrager, Roux, Dollar, Zvyagina, Ratanchandani, Yuvraj, Liang, Alao, Rodriguez, Ayub, Murthy, Nayani, Mitra, Parthasarathy, Li, Hogan, Battey, Wang, Howes, Rinott, Mehta, Siby, Bondu, Datta, Chugh, Hunt, Dhillon, Sidorov, Pan, Mahajan, Verma, Yamamoto, Ramaswamy, Lindsay, Lindsay, Feng, Lin, Zha, Patil, Shankar, Zhang, Zhang, Wang, Agarwal, Sajuyigbe, Chintala, Max, Chen, Kehoe, Satterfield, Govindaprasad, Gupta, Deng, Cho, Virk, Subramanian, Choudhury,
  Goldman, Remez, Glaser, Best, Koehler, Robinson, Li, Zhang, Matthews, Chou, Shaked, Vontimitta, Ajayi, Montanez, Mohan, Kumar, Mangla, Ionescu, Poenaru, Mihailescu, Ivanov, Li, Wang, Jiang, Bouaziz, Constable, Tang, Wu, Wang, Wu, Gao, Kleinman, Chen, Hu, Jia, Qi, Li, Zhang, Zhang, Adi, Nam, Yu, Wang, Zhao, Hao, Qian, Li, He, Rait, DeVito, Rosnbrick, Wen, Yang, Zhao, and Ma]{grattafiori2024llama3herdmodels}
Aaron Grattafiori, Abhimanyu Dubey, Abhinav Jauhri, Abhinav Pandey, Abhishek Kadian, Ahmad Al-Dahle, Aiesha Letman, Akhil Mathur, Alan Schelten, Alex Vaughan, Amy Yang, Angela Fan, Anirudh Goyal, Anthony Hartshorn, Aobo Yang, Archi Mitra, Archie Sravankumar, Artem Korenev, Arthur Hinsvark, Arun Rao, Aston Zhang, Aurelien Rodriguez, Austen Gregerson, Ava Spataru, Baptiste Roziere, Bethany Biron, Binh Tang, Bobbie Chern, Charlotte Caucheteux, Chaya Nayak, Chloe Bi, Chris Marra, Chris McConnell, Christian Keller, Christophe Touret, Chunyang Wu, Corinne Wong, Cristian~Canton Ferrer, Cyrus Nikolaidis, Damien Allonsius, Daniel Song, Danielle Pintz, Danny Livshits, Danny Wyatt, David Esiobu, Dhruv Choudhary, Dhruv Mahajan, Diego Garcia-Olano, Diego Perino, Dieuwke Hupkes, Egor Lakomkin, Ehab AlBadawy, Elina Lobanova, Emily Dinan, Eric~Michael Smith, Filip Radenovic, Francisco Guzmán, Frank Zhang, Gabriel Synnaeve, Gabrielle Lee, Georgia~Lewis Anderson, Govind Thattai, Graeme Nail, Gregoire Mialon, Guan Pang,
  Guillem Cucurell, Hailey Nguyen, Hannah Korevaar, Hu~Xu, Hugo Touvron, Iliyan Zarov, Imanol~Arrieta Ibarra, Isabel Kloumann, Ishan Misra, Ivan Evtimov, Jack Zhang, Jade Copet, Jaewon Lee, Jan Geffert, Jana Vranes, Jason Park, Jay Mahadeokar, Jeet Shah, Jelmer van~der Linde, Jennifer Billock, Jenny Hong, Jenya Lee, Jeremy Fu, Jianfeng Chi, Jianyu Huang, Jiawen Liu, Jie Wang, Jiecao Yu, Joanna Bitton, Joe Spisak, Jongsoo Park, Joseph Rocca, Joshua Johnstun, Joshua Saxe, Junteng Jia, Kalyan~Vasuden Alwala, Karthik Prasad, Kartikeya Upasani, Kate Plawiak, Ke~Li, Kenneth Heafield, Kevin Stone, Khalid El-Arini, Krithika Iyer, Kshitiz Malik, Kuenley Chiu, Kunal Bhalla, Kushal Lakhotia, Lauren Rantala-Yeary, Laurens van~der Maaten, Lawrence Chen, Liang Tan, Liz Jenkins, Louis Martin, Lovish Madaan, Lubo Malo, Lukas Blecher, Lukas Landzaat, Luke de~Oliveira, Madeline Muzzi, Mahesh Pasupuleti, Mannat Singh, Manohar Paluri, Marcin Kardas, Maria Tsimpoukelli, Mathew Oldham, Mathieu Rita, Maya Pavlova, Melanie Kambadur,
  Mike Lewis, Min Si, Mitesh~Kumar Singh, Mona Hassan, Naman Goyal, Narjes Torabi, Nikolay Bashlykov, Nikolay Bogoychev, Niladri Chatterji, Ning Zhang, Olivier Duchenne, Onur Çelebi, Patrick Alrassy, Pengchuan Zhang, Pengwei Li, Petar Vasic, Peter Weng, Prajjwal Bhargava, Pratik Dubal, Praveen Krishnan, Punit~Singh Koura, Puxin Xu, Qing He, Qingxiao Dong, Ragavan Srinivasan, Raj Ganapathy, Ramon Calderer, Ricardo~Silveira Cabral, Robert Stojnic, Roberta Raileanu, Rohan Maheswari, Rohit Girdhar, Rohit Patel, Romain Sauvestre, Ronnie Polidoro, Roshan Sumbaly, Ross Taylor, Ruan Silva, Rui Hou, Rui Wang, Saghar Hosseini, Sahana Chennabasappa, Sanjay Singh, Sean Bell, Seohyun~Sonia Kim, Sergey Edunov, Shaoliang Nie, Sharan Narang, Sharath Raparthy, Sheng Shen, Shengye Wan, Shruti Bhosale, Shun Zhang, Simon Vandenhende, Soumya Batra, Spencer Whitman, Sten Sootla, Stephane Collot, Suchin Gururangan, Sydney Borodinsky, Tamar Herman, Tara Fowler, Tarek Sheasha, Thomas Georgiou, Thomas Scialom, Tobias Speckbacher,
  Todor Mihaylov, Tong Xiao, Ujjwal Karn, Vedanuj Goswami, Vibhor Gupta, Vignesh Ramanathan, Viktor Kerkez, Vincent Gonguet, Virginie Do, Vish Vogeti, Vítor Albiero, Vladan Petrovic, Weiwei Chu, Wenhan Xiong, Wenyin Fu, Whitney Meers, Xavier Martinet, Xiaodong Wang, Xiaofang Wang, Xiaoqing~Ellen Tan, Xide Xia, Xinfeng Xie, Xuchao Jia, Xuewei Wang, Yaelle Goldschlag, Yashesh Gaur, Yasmine Babaei, Yi~Wen, Yiwen Song, Yuchen Zhang, Yue Li, Yuning Mao, Zacharie~Delpierre Coudert, Zheng Yan, Zhengxing Chen, Zoe Papakipos, Aaditya Singh, Aayushi Srivastava, Abha Jain, Adam Kelsey, Adam Shajnfeld, Adithya Gangidi, Adolfo Victoria, Ahuva Goldstand, Ajay Menon, Ajay Sharma, Alex Boesenberg, Alexei Baevski, Allie Feinstein, Amanda Kallet, Amit Sangani, Amos Teo, Anam Yunus, Andrei Lupu, Andres Alvarado, Andrew Caples, Andrew Gu, Andrew Ho, Andrew Poulton, Andrew Ryan, Ankit Ramchandani, Annie Dong, Annie Franco, Anuj Goyal, Aparajita Saraf, Arkabandhu Chowdhury, Ashley Gabriel, Ashwin Bharambe, Assaf Eisenman, Azadeh
  Yazdan, Beau James, Ben Maurer, Benjamin Leonhardi, Bernie Huang, Beth Loyd, Beto~De Paola, Bhargavi Paranjape, Bing Liu, Bo~Wu, Boyu Ni, Braden Hancock, Bram Wasti, Brandon Spence, Brani Stojkovic, Brian Gamido, Britt Montalvo, Carl Parker, Carly Burton, Catalina Mejia, Ce~Liu, Changhan Wang, Changkyu Kim, Chao Zhou, Chester Hu, Ching-Hsiang Chu, Chris Cai, Chris Tindal, Christoph Feichtenhofer, Cynthia Gao, Damon Civin, Dana Beaty, Daniel Kreymer, Daniel Li, David Adkins, David Xu, Davide Testuggine, Delia David, Devi Parikh, Diana Liskovich, Didem Foss, Dingkang Wang, Duc Le, Dustin Holland, Edward Dowling, Eissa Jamil, Elaine Montgomery, Eleonora Presani, Emily Hahn, Emily Wood, Eric-Tuan Le, Erik Brinkman, Esteban Arcaute, Evan Dunbar, Evan Smothers, Fei Sun, Felix Kreuk, Feng Tian, Filippos Kokkinos, Firat Ozgenel, Francesco Caggioni, Frank Kanayet, Frank Seide, Gabriela~Medina Florez, Gabriella Schwarz, Gada Badeer, Georgia Swee, Gil Halpern, Grant Herman, Grigory Sizov, Guangyi, Zhang, Guna
  Lakshminarayanan, Hakan Inan, Hamid Shojanazeri, Han Zou, Hannah Wang, Hanwen Zha, Haroun Habeeb, Harrison Rudolph, Helen Suk, Henry Aspegren, Hunter Goldman, Hongyuan Zhan, Ibrahim Damlaj, Igor Molybog, Igor Tufanov, Ilias Leontiadis, Irina-Elena Veliche, Itai Gat, Jake Weissman, James Geboski, James Kohli, Janice Lam, Japhet Asher, Jean-Baptiste Gaya, Jeff Marcus, Jeff Tang, Jennifer Chan, Jenny Zhen, Jeremy Reizenstein, Jeremy Teboul, Jessica Zhong, Jian Jin, Jingyi Yang, Joe Cummings, Jon Carvill, Jon Shepard, Jonathan McPhie, Jonathan Torres, Josh Ginsburg, Junjie Wang, Kai Wu, Kam~Hou U, Karan Saxena, Kartikay Khandelwal, Katayoun Zand, Kathy Matosich, Kaushik Veeraraghavan, Kelly Michelena, Keqian Li, Kiran Jagadeesh, Kun Huang, Kunal Chawla, Kyle Huang, Lailin Chen, Lakshya Garg, Lavender A, Leandro Silva, Lee Bell, Lei Zhang, Liangpeng Guo, Licheng Yu, Liron Moshkovich, Luca Wehrstedt, Madian Khabsa, Manav Avalani, Manish Bhatt, Martynas Mankus, Matan Hasson, Matthew Lennie, Matthias Reso, Maxim
  Groshev, Maxim Naumov, Maya Lathi, Meghan Keneally, Miao Liu, Michael~L. Seltzer, Michal Valko, Michelle Restrepo, Mihir Patel, Mik Vyatskov, Mikayel Samvelyan, Mike Clark, Mike Macey, Mike Wang, Miquel~Jubert Hermoso, Mo~Metanat, Mohammad Rastegari, Munish Bansal, Nandhini Santhanam, Natascha Parks, Natasha White, Navyata Bawa, Nayan Singhal, Nick Egebo, Nicolas Usunier, Nikhil Mehta, Nikolay~Pavlovich Laptev, Ning Dong, Norman Cheng, Oleg Chernoguz, Olivia Hart, Omkar Salpekar, Ozlem Kalinli, Parkin Kent, Parth Parekh, Paul Saab, Pavan Balaji, Pedro Rittner, Philip Bontrager, Pierre Roux, Piotr Dollar, Polina Zvyagina, Prashant Ratanchandani, Pritish Yuvraj, Qian Liang, Rachad Alao, Rachel Rodriguez, Rafi Ayub, Raghotham Murthy, Raghu Nayani, Rahul Mitra, Rangaprabhu Parthasarathy, Raymond Li, Rebekkah Hogan, Robin Battey, Rocky Wang, Russ Howes, Ruty Rinott, Sachin Mehta, Sachin Siby, Sai~Jayesh Bondu, Samyak Datta, Sara Chugh, Sara Hunt, Sargun Dhillon, Sasha Sidorov, Satadru Pan, Saurabh Mahajan,
  Saurabh Verma, Seiji Yamamoto, Sharadh Ramaswamy, Shaun Lindsay, Shaun Lindsay, Sheng Feng, Shenghao Lin, Shengxin~Cindy Zha, Shishir Patil, Shiva Shankar, Shuqiang Zhang, Shuqiang Zhang, Sinong Wang, Sneha Agarwal, Soji Sajuyigbe, Soumith Chintala, Stephanie Max, Stephen Chen, Steve Kehoe, Steve Satterfield, Sudarshan Govindaprasad, Sumit Gupta, Summer Deng, Sungmin Cho, Sunny Virk, Suraj Subramanian, Sy~Choudhury, Sydney Goldman, Tal Remez, Tamar Glaser, Tamara Best, Thilo Koehler, Thomas Robinson, Tianhe Li, Tianjun Zhang, Tim Matthews, Timothy Chou, Tzook Shaked, Varun Vontimitta, Victoria Ajayi, Victoria Montanez, Vijai Mohan, Vinay~Satish Kumar, Vishal Mangla, Vlad Ionescu, Vlad Poenaru, Vlad~Tiberiu Mihailescu, Vladimir Ivanov, Wei Li, Wenchen Wang, Wenwen Jiang, Wes Bouaziz, Will Constable, Xiaocheng Tang, Xiaojian Wu, Xiaolan Wang, Xilun Wu, Xinbo Gao, Yaniv Kleinman, Yanjun Chen, Ye~Hu, Ye~Jia, Ye~Qi, Yenda Li, Yilin Zhang, Ying Zhang, Yossi Adi, Youngjin Nam, Yu, Wang, Yu~Zhao, Yuchen Hao, Yundi
  Qian, Yunlu Li, Yuzi He, Zach Rait, Zachary DeVito, Zef Rosnbrick, Zhaoduo Wen, Zhenyu Yang, Zhiwei Zhao, and Zhiyu Ma.
\newblock The llama 3 herd of models, 2024.
\newblock URL \url{https://arxiv.org/abs/2407.21783}.

\bibitem[Guo et~al.(2025)Guo, Yang, Zhang, Song, Zhang, Xu, Zhu, Ma, Wang, Bi, et~al.]{guo2025deepseek}
Daya Guo, Dejian Yang, Haowei Zhang, Junxiao Song, Ruoyu Zhang, Runxin Xu, Qihao Zhu, Shirong Ma, Peiyi Wang, Xiao Bi, et~al.
\newblock Deepseek-r1: Incentivizing reasoning capability in llms via reinforcement learning.
\newblock \emph{arXiv preprint arXiv:2501.12948}, 2025.

\bibitem[Hao et~al.(2023)Hao, Gu, Ma, Hong, Wang, Wang, and Hu]{hao2023reasoninglanguagemodelplanning}
Shibo Hao, Yi~Gu, Haodi Ma, Joshua~Jiahua Hong, Zhen Wang, Daisy~Zhe Wang, and Zhiting Hu.
\newblock Reasoning with language model is planning with world model, 2023.
\newblock URL \url{https://arxiv.org/abs/2305.14992}.

\bibitem[He et~al.(2024)He, Luo, Bai, Hu, Thai, Shen, Hu, Han, Huang, Zhang, et~al.]{he2024olympiadbench}
Chaoqun He, Renjie Luo, Yuzhuo Bai, Shengding Hu, Zhen~Leng Thai, Junhao Shen, Jinyi Hu, Xu~Han, Yujie Huang, Yuxiang Zhang, et~al.
\newblock Olympiadbench: A challenging benchmark for promoting agi with olympiad-level bilingual multimodal scientific problems.
\newblock \emph{arXiv preprint arXiv:2402.14008}, 2024.

\bibitem[Hendrycks et~al.(2021)Hendrycks, Burns, Kadavath, Arora, Basart, Tang, Song, and Steinhardt]{hendrycks2021measuring}
Dan Hendrycks, Collin Burns, Saurav Kadavath, Akul Arora, Steven Basart, Eric Tang, Dawn Song, and Jacob Steinhardt.
\newblock Measuring mathematical problem solving with the math dataset.
\newblock \emph{arXiv preprint arXiv:2103.03874}, 2021.

\bibitem[Huang et~al.(2024)Huang, Zou, Li, Liu, Zheng, Chern, Xia, Qin, Yuan, and Liu]{huang2024o1}
Zhen Huang, Haoyang Zou, Xuefeng Li, Yixiu Liu, Yuxiang Zheng, Ethan Chern, Shijie Xia, Yiwei Qin, Weizhe Yuan, and Pengfei Liu.
\newblock O1 replication journey--part 2: Surpassing o1-preview through simple distillation, big progress or bitter lesson?
\newblock \emph{arXiv preprint arXiv:2411.16489}, 2024.

\bibitem[Kambhampati(2024)]{Kambhampati_2024}
Subbarao Kambhampati.
\newblock Can large language models reason and plan?
\newblock \emph{Annals of the New York Academy of Sciences}, 1534\penalty0 (1):\penalty0 15–18, March 2024.
\newblock ISSN 1749-6632.
\newblock \doi{10.1111/nyas.15125}.
\newblock URL \url{http://dx.doi.org/10.1111/nyas.15125}.

\bibitem[Kaplan et~al.(2020)Kaplan, McCandlish, Henighan, Brown, Chess, Child, Gray, Radford, Wu, and Amodei]{kaplan2020scalinglawsneurallanguage}
Jared Kaplan, Sam McCandlish, Tom Henighan, Tom~B. Brown, Benjamin Chess, Rewon Child, Scott Gray, Alec Radford, Jeffrey Wu, and Dario Amodei.
\newblock Scaling laws for neural language models, 2020.
\newblock URL \url{https://arxiv.org/abs/2001.08361}.

\bibitem[Lewkowycz et~al.(2022)Lewkowycz, Andreassen, Dohan, Dyer, Michalewski, Ramasesh, Slone, Anil, Schlag, Gutman-Solo, et~al.]{lewkowycz2022solving}
Aitor Lewkowycz, Anders Andreassen, David Dohan, Ethan Dyer, Henryk Michalewski, Vinay Ramasesh, Ambrose Slone, Cem Anil, Imanol Schlag, Theo Gutman-Solo, et~al.
\newblock Solving quantitative reasoning problems with language models.
\newblock \emph{Advances in Neural Information Processing Systems}, 35:\penalty0 3843--3857, 2022.

\bibitem[Li et~al.(2024{\natexlab{a}})Li, Wang, Hu, Wei, Zheng, Hu, Zhang, and Peng]{li2024common7blanguagemodels}
Chen Li, Weiqi Wang, Jingcheng Hu, Yixuan Wei, Nanning Zheng, Han Hu, Zheng Zhang, and Houwen Peng.
\newblock Common 7b language models already possess strong math capabilities, 2024{\natexlab{a}}.
\newblock URL \url{https://arxiv.org/abs/2403.04706}.

\bibitem[Li et~al.(2024{\natexlab{b}})Li, Beeching, Tunstall, Lipkin, Soletskyi, Huang, Rasul, Yu, Jiang, Shen, et~al.]{li2024numinamath}
Jia Li, Edward Beeching, Lewis Tunstall, Ben Lipkin, Roman Soletskyi, Shengyi Huang, Kashif Rasul, Longhui Yu, Albert~Q Jiang, Ziju Shen, et~al.
\newblock Numinamath: The largest public dataset in ai4maths with 860k pairs of competition math problems and solutions.
\newblock \emph{Hugging Face repository}, 2024{\natexlab{b}}.

\bibitem[Li et~al.(2022)Li, Choi, Chung, Kushman, Schrittwieser, Leblond, Eccles, Keeling, Gimeno, Dal~Lago, Hubert, Choy, de~Masson~d’Autume, Babuschkin, Chen, Huang, Welbl, Gowal, Cherepanov, Molloy, Mankowitz, Sutherland~Robson, Kohli, de~Freitas, Kavukcuoglu, and Vinyals]{Li_2022}
Yujia Li, David Choi, Junyoung Chung, Nate Kushman, Julian Schrittwieser, Rémi Leblond, Tom Eccles, James Keeling, Felix Gimeno, Agustin Dal~Lago, Thomas Hubert, Peter Choy, Cyprien de~Masson~d’Autume, Igor Babuschkin, Xinyun Chen, Po-Sen Huang, Johannes Welbl, Sven Gowal, Alexey Cherepanov, James Molloy, Daniel~J. Mankowitz, Esme Sutherland~Robson, Pushmeet Kohli, Nando de~Freitas, Koray Kavukcuoglu, and Oriol Vinyals.
\newblock Competition-level code generation with alphacode.
\newblock \emph{Science}, 378\penalty0 (6624):\penalty0 1092–1097, December 2022.
\newblock ISSN 1095-9203.
\newblock \doi{10.1126/science.abq1158}.
\newblock URL \url{http://dx.doi.org/10.1126/science.abq1158}.

\bibitem[Luo et~al.(2025)Luo, Tan, Wong, Shi, Tang, Roongta, Cai, Luo, Zhang, Li, Popa, and Stoica]{deepscaler2025}
Michael Luo, Sijun Tan, Justin Wong, Xiaoxiang Shi, William Tang, Manan Roongta, Colin Cai, Jeffrey Luo, Tianjun Zhang, Erran Li, Raluca~Ada Popa, and Ion Stoica.
\newblock Deepscaler: Surpassing o1-preview with a 1.5b model by scaling rl.
\newblock \url{https://tinyurl.com/DeepScaleR}, 2025.
\newblock Notion Blog.

\bibitem[Merrill \& Sabharwal(2024)Merrill and Sabharwal]{merrill2024expressivepowertransformerschain}
William Merrill and Ashish Sabharwal.
\newblock The expressive power of transformers with chain of thought, 2024.
\newblock URL \url{https://arxiv.org/abs/2310.07923}.

\bibitem[Mirzadeh et~al.(2024)Mirzadeh, Alizadeh, Shahrokhi, Tuzel, Bengio, and Farajtabar]{mirzadeh2024gsmsymbolicunderstandinglimitationsmathematical}
Iman Mirzadeh, Keivan Alizadeh, Hooman Shahrokhi, Oncel Tuzel, Samy Bengio, and Mehrdad Farajtabar.
\newblock Gsm-symbolic: Understanding the limitations of mathematical reasoning in large language models, 2024.
\newblock URL \url{https://arxiv.org/abs/2410.05229}.

\bibitem[OpenAI(2024)]{openai-o1}
OpenAI.
\newblock Learning to reason with llms, september 2024, 2024.
\newblock URL \url{https://openai.com/index/learning-to-reason-with-llms/}.

\bibitem[OpenAI et~al.(2024)OpenAI, :, Jaech, Kalai, Lerer, Richardson, El-Kishky, Low, Helyar, Madry, Beutel, Carney, Iftimie, Karpenko, Passos, Neitz, Prokofiev, Wei, Tam, Bennett, Kumar, Saraiva, Vallone, Duberstein, Kondrich, Mishchenko, Applebaum, Jiang, Nair, Zoph, Ghorbani, Rossen, Sokolowsky, Barak, McGrew, Minaiev, Hao, Baker, Houghton, McKinzie, Eastman, Lugaresi, Bassin, Hudson, Li, de~Bourcy, Voss, Shen, Zhang, Koch, Orsinger, Hesse, Fischer, Chan, Roberts, Kappler, Levy, Selsam, Dohan, Farhi, Mely, Robinson, Tsipras, Li, Oprica, Freeman, Zhang, Wong, Proehl, Cheung, Mitchell, Wallace, Ritter, Mays, Wang, Such, Raso, Leoni, Tsimpourlas, Song, von Lohmann, Sulit, Salmon, Parascandolo, Chabot, Zhao, Brockman, Leclerc, Salman, Bao, Sheng, Andrin, Bagherinezhad, Ren, Lightman, Chung, Kivlichan, O'Connell, Osband, Gilaberte, Akkaya, Kostrikov, Sutskever, Kofman, Pachocki, Lennon, Wei, Harb, Twore, Feng, Yu, Weng, Tang, Yu, Candela, Palermo, Parish, Heidecke, Hallman, Rizzo, Gordon, Uesato, Ward,
  Huizinga, Wang, Chen, Xiao, Singhal, Nguyen, Cobbe, Shi, Wood, Rimbach, Gu-Lemberg, Liu, Lu, Stone, Yu, Ahmad, Yang, Liu, Maksin, Ho, Fedus, Weng, Li, McCallum, Held, Kuhn, Kondraciuk, Kaiser, Metz, Boyd, Trebacz, Joglekar, Chen, Tintor, Meyer, Jones, Kaufer, Schwarzer, Shah, Yatbaz, Guan, Xu, Yan, Glaese, Chen, Lampe, Malek, Wang, Fradin, McClay, Pavlov, Wang, Wang, Murati, Bavarian, Rohaninejad, McAleese, Chowdhury, Chowdhury, Ryder, Tezak, Brown, Nachum, Boiko, Murk, Watkins, Chao, Ashbourne, Izmailov, Zhokhov, Dias, Arora, Lin, Lopes, Gaon, Miyara, Leike, Hwang, Garg, Brown, James, Shu, Cheu, Greene, Jain, Altman, Toizer, Toyer, Miserendino, Agarwal, Hernandez, Baker, McKinney, Yan, Zhao, Hu, Santurkar, Chaudhuri, Zhang, Fu, Papay, Lin, Balaji, Sanjeev, Sidor, Broda, Clark, Wang, Gordon, Sanders, Patwardhan, Sottiaux, Degry, Dimson, Zheng, Garipov, Stasi, Bansal, Creech, Peterson, Eloundou, Qi, Kosaraju, Monaco, Pong, Fomenko, Zheng, Zhou, McCabe, Zaremba, Dubois, Lu, Chen, Cha, Bai, He, Zhang, Wang,
  Shao, and Li]{openai2024openaio1card}
OpenAI, :, Aaron Jaech, Adam Kalai, Adam Lerer, Adam Richardson, Ahmed El-Kishky, Aiden Low, Alec Helyar, Aleksander Madry, Alex Beutel, Alex Carney, Alex Iftimie, Alex Karpenko, Alex~Tachard Passos, Alexander Neitz, Alexander Prokofiev, Alexander Wei, Allison Tam, Ally Bennett, Ananya Kumar, Andre Saraiva, Andrea Vallone, Andrew Duberstein, Andrew Kondrich, Andrey Mishchenko, Andy Applebaum, Angela Jiang, Ashvin Nair, Barret Zoph, Behrooz Ghorbani, Ben Rossen, Benjamin Sokolowsky, Boaz Barak, Bob McGrew, Borys Minaiev, Botao Hao, Bowen Baker, Brandon Houghton, Brandon McKinzie, Brydon Eastman, Camillo Lugaresi, Cary Bassin, Cary Hudson, Chak~Ming Li, Charles de~Bourcy, Chelsea Voss, Chen Shen, Chong Zhang, Chris Koch, Chris Orsinger, Christopher Hesse, Claudia Fischer, Clive Chan, Dan Roberts, Daniel Kappler, Daniel Levy, Daniel Selsam, David Dohan, David Farhi, David Mely, David Robinson, Dimitris Tsipras, Doug Li, Dragos Oprica, Eben Freeman, Eddie Zhang, Edmund Wong, Elizabeth Proehl, Enoch Cheung, Eric
  Mitchell, Eric Wallace, Erik Ritter, Evan Mays, Fan Wang, Felipe~Petroski Such, Filippo Raso, Florencia Leoni, Foivos Tsimpourlas, Francis Song, Fred von Lohmann, Freddie Sulit, Geoff Salmon, Giambattista Parascandolo, Gildas Chabot, Grace Zhao, Greg Brockman, Guillaume Leclerc, Hadi Salman, Haiming Bao, Hao Sheng, Hart Andrin, Hessam Bagherinezhad, Hongyu Ren, Hunter Lightman, Hyung~Won Chung, Ian Kivlichan, Ian O'Connell, Ian Osband, Ignasi~Clavera Gilaberte, Ilge Akkaya, Ilya Kostrikov, Ilya Sutskever, Irina Kofman, Jakub Pachocki, James Lennon, Jason Wei, Jean Harb, Jerry Twore, Jiacheng Feng, Jiahui Yu, Jiayi Weng, Jie Tang, Jieqi Yu, Joaquin~Quiñonero Candela, Joe Palermo, Joel Parish, Johannes Heidecke, John Hallman, John Rizzo, Jonathan Gordon, Jonathan Uesato, Jonathan Ward, Joost Huizinga, Julie Wang, Kai Chen, Kai Xiao, Karan Singhal, Karina Nguyen, Karl Cobbe, Katy Shi, Kayla Wood, Kendra Rimbach, Keren Gu-Lemberg, Kevin Liu, Kevin Lu, Kevin Stone, Kevin Yu, Lama Ahmad, Lauren Yang, Leo Liu,
  Leon Maksin, Leyton Ho, Liam Fedus, Lilian Weng, Linden Li, Lindsay McCallum, Lindsey Held, Lorenz Kuhn, Lukas Kondraciuk, Lukasz Kaiser, Luke Metz, Madelaine Boyd, Maja Trebacz, Manas Joglekar, Mark Chen, Marko Tintor, Mason Meyer, Matt Jones, Matt Kaufer, Max Schwarzer, Meghan Shah, Mehmet Yatbaz, Melody~Y. Guan, Mengyuan Xu, Mengyuan Yan, Mia Glaese, Mianna Chen, Michael Lampe, Michael Malek, Michele Wang, Michelle Fradin, Mike McClay, Mikhail Pavlov, Miles Wang, Mingxuan Wang, Mira Murati, Mo~Bavarian, Mostafa Rohaninejad, Nat McAleese, Neil Chowdhury, Neil Chowdhury, Nick Ryder, Nikolas Tezak, Noam Brown, Ofir Nachum, Oleg Boiko, Oleg Murk, Olivia Watkins, Patrick Chao, Paul Ashbourne, Pavel Izmailov, Peter Zhokhov, Rachel Dias, Rahul Arora, Randall Lin, Rapha~Gontijo Lopes, Raz Gaon, Reah Miyara, Reimar Leike, Renny Hwang, Rhythm Garg, Robin Brown, Roshan James, Rui Shu, Ryan Cheu, Ryan Greene, Saachi Jain, Sam Altman, Sam Toizer, Sam Toyer, Samuel Miserendino, Sandhini Agarwal, Santiago Hernandez,
  Sasha Baker, Scott McKinney, Scottie Yan, Shengjia Zhao, Shengli Hu, Shibani Santurkar, Shraman~Ray Chaudhuri, Shuyuan Zhang, Siyuan Fu, Spencer Papay, Steph Lin, Suchir Balaji, Suvansh Sanjeev, Szymon Sidor, Tal Broda, Aidan Clark, Tao Wang, Taylor Gordon, Ted Sanders, Tejal Patwardhan, Thibault Sottiaux, Thomas Degry, Thomas Dimson, Tianhao Zheng, Timur Garipov, Tom Stasi, Trapit Bansal, Trevor Creech, Troy Peterson, Tyna Eloundou, Valerie Qi, Vineet Kosaraju, Vinnie Monaco, Vitchyr Pong, Vlad Fomenko, Weiyi Zheng, Wenda Zhou, Wes McCabe, Wojciech Zaremba, Yann Dubois, Yinghai Lu, Yining Chen, Young Cha, Yu~Bai, Yuchen He, Yuchen Zhang, Yunyun Wang, Zheng Shao, and Zhuohan Li.
\newblock Openai o1 system card, 2024.
\newblock URL \url{https://arxiv.org/abs/2412.16720}.

\bibitem[Paster et~al.(2023)Paster, Santos, Azerbayev, and Ba]{paster2023openwebmathopendatasethighquality}
Keiran Paster, Marco~Dos Santos, Zhangir Azerbayev, and Jimmy Ba.
\newblock Openwebmath: An open dataset of high-quality mathematical web text, 2023.
\newblock URL \url{https://arxiv.org/abs/2310.06786}.

\bibitem[Qin et~al.(2024)Qin, Li, Zou, Liu, Xia, Huang, Ye, Yuan, Liu, Li, et~al.]{qin2024o1}
Yiwei Qin, Xuefeng Li, Haoyang Zou, Yixiu Liu, Shijie Xia, Zhen Huang, Yixin Ye, Weizhe Yuan, Hector Liu, Yuanzhi Li, et~al.
\newblock O1 replication journey: A strategic progress report--part 1.
\newblock \emph{arXiv preprint arXiv:2410.18982}, 2024.

\bibitem[Qwen et~al.(2025)Qwen, :, Yang, Yang, Zhang, Hui, Zheng, Yu, Li, Liu, Huang, Wei, Lin, Yang, Tu, Zhang, Yang, Yang, Zhou, Lin, Dang, Lu, Bao, Yang, Yu, Li, Xue, Zhang, Zhu, Men, Lin, Li, Tang, Xia, Ren, Ren, Fan, Su, Zhang, Wan, Liu, Cui, Zhang, and Qiu]{qwen2025qwen25technicalreport}
Qwen, :, An~Yang, Baosong Yang, Beichen Zhang, Binyuan Hui, Bo~Zheng, Bowen Yu, Chengyuan Li, Dayiheng Liu, Fei Huang, Haoran Wei, Huan Lin, Jian Yang, Jianhong Tu, Jianwei Zhang, Jianxin Yang, Jiaxi Yang, Jingren Zhou, Junyang Lin, Kai Dang, Keming Lu, Keqin Bao, Kexin Yang, Le~Yu, Mei Li, Mingfeng Xue, Pei Zhang, Qin Zhu, Rui Men, Runji Lin, Tianhao Li, Tianyi Tang, Tingyu Xia, Xingzhang Ren, Xuancheng Ren, Yang Fan, Yang Su, Yichang Zhang, Yu~Wan, Yuqiong Liu, Zeyu Cui, Zhenru Zhang, and Zihan Qiu.
\newblock Qwen2.5 technical report, 2025.
\newblock URL \url{https://arxiv.org/abs/2412.15115}.

\bibitem[Rajbhandari et~al.(2020)Rajbhandari, Rasley, Ruwase, and He]{rajbhandari2020zero}
Samyam Rajbhandari, Jeff Rasley, Olatunji Ruwase, and Yuxiong He.
\newblock Zero: Memory optimizations toward training trillion parameter models.
\newblock In \emph{SC20: International Conference for High Performance Computing, Networking, Storage and Analysis}, pp.\  1--16. IEEE, 2020.

\bibitem[Rein et~al.(2023)Rein, Hou, Stickland, Petty, Pang, Dirani, Michael, and Bowman]{rein2023gpqa}
David Rein, Betty~Li Hou, Asa~Cooper Stickland, Jackson Petty, Richard~Yuanzhe Pang, Julien Dirani, Julian Michael, and Samuel~R Bowman.
\newblock Gpqa: A graduate-level google-proof q\&a benchmark.
\newblock \emph{arXiv preprint arXiv:2311.12022}, 2023.

\bibitem[Shao et~al.(2024)Shao, Wang, Zhu, Xu, Song, Zhang, Li, Wu, and Guo]{shao2024deepseekmath}
Zhihong Shao, Peiyi Wang, Qihao Zhu, Runxin Xu, Junxiao Song, Mingchuan Zhang, YK~Li, Yu~Wu, and Daya Guo.
\newblock Deepseekmath: Pushing the limits of mathematical reasoning in open language models.
\newblock \emph{arXiv preprint arXiv:2402.03300}, 2024.

\bibitem[Snell et~al.(2024)Snell, Lee, Xu, and Kumar]{snell2024scaling}
Charlie Snell, Jaehoon Lee, Kelvin Xu, and Aviral Kumar.
\newblock Scaling llm test-time compute optimally can be more effective than scaling model parameters.
\newblock \emph{arXiv preprint arXiv:2408.03314}, 2024.

\bibitem[Team(2025{\natexlab{a}})]{openthoughts}
OpenThoughts Team.
\newblock {Open Thoughts}.
\newblock https://open-thoughts.ai, January 2025{\natexlab{a}}.

\bibitem[Team(2024{\natexlab{a}})]{qwen1.5}
Qwen Team.
\newblock Introducing qwen1.5, 2024{\natexlab{a}}.
\newblock URL \url{https://qwenlm.github.io/blog/qwen1.5/}.

\bibitem[Team(2024{\natexlab{b}})]{qwq}
Qwen Team.
\newblock Qwq: Reflect deeply on the boundaries of the unknown, 2024{\natexlab{b}}.
\newblock URL \url{https://qwenlm.github.io/blog/qwq-32b-preview/}.

\bibitem[Team(2025{\natexlab{b}})]{qwq32b}
Qwen Team.
\newblock Qwq-32b: Embracing the power of reinforcement learning, March 2025{\natexlab{b}}.
\newblock URL \url{https://qwenlm.github.io/blog/qwq-32b/}.

\bibitem[Touvron et~al.(2023)Touvron, Martin, Stone, Albert, Almahairi, Babaei, Bashlykov, Batra, Bhargava, Bhosale, Bikel, Blecher, Ferrer, Chen, Cucurull, Esiobu, Fernandes, Fu, Fu, Fuller, Gao, Goswami, Goyal, Hartshorn, Hosseini, Hou, Inan, Kardas, Kerkez, Khabsa, Kloumann, Korenev, Koura, Lachaux, Lavril, Lee, Liskovich, Lu, Mao, Martinet, Mihaylov, Mishra, Molybog, Nie, Poulton, Reizenstein, Rungta, Saladi, Schelten, Silva, Smith, Subramanian, Tan, Tang, Taylor, Williams, Kuan, Xu, Yan, Zarov, Zhang, Fan, Kambadur, Narang, Rodriguez, Stojnic, Edunov, and Scialom]{touvron2023llama2openfoundation}
Hugo Touvron, Louis Martin, Kevin Stone, Peter Albert, Amjad Almahairi, Yasmine Babaei, Nikolay Bashlykov, Soumya Batra, Prajjwal Bhargava, Shruti Bhosale, Dan Bikel, Lukas Blecher, Cristian~Canton Ferrer, Moya Chen, Guillem Cucurull, David Esiobu, Jude Fernandes, Jeremy Fu, Wenyin Fu, Brian Fuller, Cynthia Gao, Vedanuj Goswami, Naman Goyal, Anthony Hartshorn, Saghar Hosseini, Rui Hou, Hakan Inan, Marcin Kardas, Viktor Kerkez, Madian Khabsa, Isabel Kloumann, Artem Korenev, Punit~Singh Koura, Marie-Anne Lachaux, Thibaut Lavril, Jenya Lee, Diana Liskovich, Yinghai Lu, Yuning Mao, Xavier Martinet, Todor Mihaylov, Pushkar Mishra, Igor Molybog, Yixin Nie, Andrew Poulton, Jeremy Reizenstein, Rashi Rungta, Kalyan Saladi, Alan Schelten, Ruan Silva, Eric~Michael Smith, Ranjan Subramanian, Xiaoqing~Ellen Tan, Binh Tang, Ross Taylor, Adina Williams, Jian~Xiang Kuan, Puxin Xu, Zheng Yan, Iliyan Zarov, Yuchen Zhang, Angela Fan, Melanie Kambadur, Sharan Narang, Aurelien Rodriguez, Robert Stojnic, Sergey Edunov, and Thomas
  Scialom.
\newblock Llama 2: Open foundation and fine-tuned chat models, 2023.
\newblock URL \url{https://arxiv.org/abs/2307.09288}.

\bibitem[Wang et~al.(2022)Wang, Wei, Schuurmans, Le, Chi, Narang, Chowdhery, and Zhou]{wang2022self}
Xuezhi Wang, Jason Wei, Dale Schuurmans, Quoc Le, Ed~Chi, Sharan Narang, Aakanksha Chowdhery, and Denny Zhou.
\newblock Self-consistency improves chain of thought reasoning in language models.
\newblock \emph{arXiv preprint arXiv:2203.11171}, 2022.

\bibitem[Wang et~al.(2024)Wang, Li, Xia, and Liu]{wang2024mathpilebilliontokenscalepretrainingcorpus}
Zengzhi Wang, Xuefeng Li, Rui Xia, and Pengfei Liu.
\newblock Mathpile: A billion-token-scale pretraining corpus for math, 2024.
\newblock URL \url{https://arxiv.org/abs/2312.17120}.

\bibitem[Wei et~al.(2022)Wei, Wang, Schuurmans, Bosma, Xia, Chi, Le, Zhou, et~al.]{wei2022chain}
Jason Wei, Xuezhi Wang, Dale Schuurmans, Maarten Bosma, Fei Xia, Ed~Chi, Quoc~V Le, Denny Zhou, et~al.
\newblock Chain-of-thought prompting elicits reasoning in large language models.
\newblock \emph{Advances in neural information processing systems}, 35:\penalty0 24824--24837, 2022.

\bibitem[Xiang et~al.(2025)Xiang, Snell, Gandhi, Albalak, Singh, Blagden, Phung, Rafailov, Lile, Mahan, Castricato, Franken, Haber, and Finn]{xiang20252reasoningllmslearning}
Violet Xiang, Charlie Snell, Kanishk Gandhi, Alon Albalak, Anikait Singh, Chase Blagden, Duy Phung, Rafael Rafailov, Nathan Lile, Dakota Mahan, Louis Castricato, Jan-Philipp Franken, Nick Haber, and Chelsea Finn.
\newblock Towards system 2 reasoning in llms: Learning how to think with meta chain-of-thought, 2025.
\newblock URL \url{https://arxiv.org/abs/2501.04682}.

\bibitem[Xu et~al.(2024)Xu, Wang, Fan, and Liu]{xu2024benchmarkingbenchmarkleakagelarge}
Ruijie Xu, Zengzhi Wang, Run-Ze Fan, and Pengfei Liu.
\newblock Benchmarking benchmark leakage in large language models, 2024.
\newblock URL \url{https://arxiv.org/abs/2404.18824}.

\bibitem[Yang et~al.(2024)Yang, Zhang, Hui, Gao, Yu, Li, Liu, Tu, Zhou, Lin, et~al.]{yang2024qwen2}
An~Yang, Beichen Zhang, Binyuan Hui, Bofei Gao, Bowen Yu, Chengpeng Li, Dayiheng Liu, Jianhong Tu, Jingren Zhou, Junyang Lin, et~al.
\newblock Qwen2. 5-math technical report: Toward mathematical expert model via self-improvement.
\newblock \emph{arXiv preprint arXiv:2409.12122}, 2024.

\bibitem[Yao et~al.(2023)Yao, Yu, Zhao, Shafran, Griffiths, Cao, and Narasimhan]{yao2023treethoughtsdeliberateproblem}
Shunyu Yao, Dian Yu, Jeffrey Zhao, Izhak Shafran, Thomas~L. Griffiths, Yuan Cao, and Karthik Narasimhan.
\newblock Tree of thoughts: Deliberate problem solving with large language models, 2023.
\newblock URL \url{https://arxiv.org/abs/2305.10601}.

\bibitem[Yu et~al.(2024)Yu, Jiang, Shi, Yu, Liu, Zhang, Kwok, Li, Weller, and Liu]{yu2024metamathbootstrapmathematicalquestions}
Longhui Yu, Weisen Jiang, Han Shi, Jincheng Yu, Zhengying Liu, Yu~Zhang, James~T. Kwok, Zhenguo Li, Adrian Weller, and Weiyang Liu.
\newblock Metamath: Bootstrap your own mathematical questions for large language models, 2024.
\newblock URL \url{https://arxiv.org/abs/2309.12284}.

\bibitem[Yue et~al.(2023)Yue, Qu, Zhang, Fu, Huang, Sun, Su, and Chen]{yue2023mammothbuildingmathgeneralist}
Xiang Yue, Xingwei Qu, Ge~Zhang, Yao Fu, Wenhao Huang, Huan Sun, Yu~Su, and Wenhu Chen.
\newblock Mammoth: Building math generalist models through hybrid instruction tuning, 2023.
\newblock URL \url{https://arxiv.org/abs/2309.05653}.

\bibitem[Yue et~al.(2024)Yue, Zheng, Zhang, and Chen]{yue2024mammoth2scalinginstructionsweb}
Xiang Yue, Tuney Zheng, Ge~Zhang, and Wenhu Chen.
\newblock Mammoth2: Scaling instructions from the web, 2024.
\newblock URL \url{https://arxiv.org/abs/2405.03548}.

\bibitem[Zhang et~al.(2024)Zhang, Da, Lee, Robinson, Wu, Song, Zhao, Raja, Zhuang, Slack, Lyu, Hendryx, Kaplan, Lunati, and Yue]{zhang2024carefulexaminationlargelanguage}
Hugh Zhang, Jeff Da, Dean Lee, Vaughn Robinson, Catherine Wu, Will Song, Tiffany Zhao, Pranav Raja, Charlotte Zhuang, Dylan Slack, Qin Lyu, Sean Hendryx, Russell Kaplan, Michele Lunati, and Summer Yue.
\newblock A careful examination of large language model performance on grade school arithmetic, 2024.
\newblock URL \url{https://arxiv.org/abs/2405.00332}.

\bibitem[Zhou et~al.(2023)Zhou, Liu, Xu, Iyer, Sun, Mao, Ma, Efrat, Yu, Yu, et~al.]{zhou2023lima}
Chunting Zhou, Pengfei Liu, Puxin Xu, Srinivasan Iyer, Jiao Sun, Yuning Mao, Xuezhe Ma, Avia Efrat, Ping Yu, Lili Yu, et~al.
\newblock Lima: Less is more for alignment.
\newblock \emph{Advances in Neural Information Processing Systems}, 36:\penalty0 55006--55021, 2023.

\bibitem[Zhou et~al.(2024)Zhou, Wang, Liu, Li, and Liu]{zhou2024programmingexampleliftingpretraining}
Fan Zhou, Zengzhi Wang, Qian Liu, Junlong Li, and Pengfei Liu.
\newblock Programming every example: Lifting pre-training data quality like experts at scale, 2024.
\newblock URL \url{https://arxiv.org/abs/2409.17115}.

\end{thebibliography}
